\documentclass[acmsmall]{acmart}
\usepackage{algorithm}
\usepackage{algorithmic}
\usepackage{graphicx}
\usepackage{subfigure}
\usepackage{epstopdf}
\usepackage{xspace}
\usepackage{wrapfig}
\usepackage{amsthm}
\usepackage{color}
\usepackage[misc]{ifsym}
\usepackage{epstopdf}
\newtheorem{theorem}{Theorem}
\AtBeginDocument{%
  \providecommand\BibTeX{{%
    \normalfont B\kern-0.5em{\scshape i\kern-0.25em b}\kern-0.8em\TeX}}}

\setcopyright{acmcopyright}
\copyrightyear{2021}
\acmYear{2021}
\acmDOI{00.0000/0000000.0000000}

\acmJournal{TIST}
\acmVolume{0}
\acmNumber{0}
\acmArticle{000}
\acmMonth{5}

\newcommand{\method}{AFFAR\xspace}
\begin{document}

\title{Domain Generalization for Activity Recognition via Adaptive Feature Fusion}

\author{Xin Qin}
\email{qinxin18b@ict.ac.cn}
\affiliation{%
  \institution{Beijing Key Lab. of Mobile Computing and Pervasive Devices, Inst. of Computing Tech., CAS, University of Chinese Academy of Sciences}
  \city{Beijing}
  \country{China}
}

\author{Jindong Wang}
\email{jindong.wang@microsoft.com}
\authornotemark[1]
\affiliation{%
  \institution{Microsoft Research Asia}
  \city{Beijing}
  \country{China}
}

\author{Yiqiang Chen}
\authornote{Correspondence to: J. Wang and Y. Chen.}
\email{yqchen@ict.ac.cn}
\affiliation{%
  \institution{Beijing Key Lab. of Mobile Computing and Pervasive Devices, Inst. of Computing Tech., CAS, University of Chinese Academy of Sciences, Pengcheng Laboratory, Shenzhen}
  \country{China}
  }

\author{Wang Lu}
\email{luwang@ict.ac.cn}
\affiliation{%
  \institution{Beijing Key Lab. of Mobile Computing and Pervasive Devices, Inst. of Computing Tech., CAS, University of Chinese Academy of Sciences}
  \city{Beijing}
  \country{China}
}

\author{Xinlong Jiang}
\email{jiangxinlong@ict.ac.cn}
\affiliation{%
  \institution{Beijing Key Lab. of Mobile Computing and Pervasive Devices, Inst. of Computing Tech., CAS}
  \city{Beijing}
  \country{China}
  }
  
\renewcommand{\shortauthors}{Qin, et al.}

\begin{abstract}
Human activity recognition requires the efforts to build a generalizable model using the training datasets with the hope to achieve good performance in test datasets. However, in real applications, the training and testing datasets may have totally different distributions due to various reasons such as different body shapes, acting styles, and habits, damaging the model's generalization performance. While such a distribution gap can be reduced by existing domain adaptation approaches, they typically assume that the test data can be accessed in the training stage, which is not realistic. In this paper, we consider a more practical and challenging scenario: domain-generalized activity recognition (DGAR) where the test dataset \emph{cannot} be accessed during training. To this end, we propose \emph{Adaptive Feature Fusion for Activity Recognition~(AFFAR)}, a domain generalization approach that learns to fuse the domain-invariant and domain-specific representations to improve the model's generalization performance. AFFAR takes the best of both worlds where domain-invariant representations enhance the transferability across domains and domain-specific representations leverage the model discrimination power from each domain. Extensive experiments on three public HAR datasets show its effectiveness. Furthermore, we apply AFFAR to a real application, i.e., the diagnosis of Children's Attention Deficit Hyperactivity Disorder~(ADHD), which also demonstrates the superiority of our approach.
\end{abstract}

\ccsdesc[500]{Human-centered computing~Ubiquitous computing}
\ccsdesc[500]{Computing methodologies~Transfer learning}

\keywords{Human Activity Recognition, Domain Generalization, Transfer Learning}

\maketitle

\section{Introduction}
\label{sec-int}
Human activity recognition~(HAR) is an active research topic in ubiquitous computing.
HAR aims at recognizing people's activities by building machine learning models on the activity data.
HAR has been wildly applied in smart-home~\cite{feuz2014heterogeneous}, fatigue detection~\cite{sikander2018driver}, fall detection for the elder~\cite{lu2018deep}, attention deficit hyperactivity disorder (ADHD)~\cite{demontis2019discovery}, and other fields.
Therefore, accurate HAR is of vital importance in real-world applications.
Many machine learning methods have been adopted to improve the performance of HAR, such as Support Vector Machines (SVM), K-Nearest Neighbor (KNN), Random Forest~(RF), and the deep learning models including Convolutional Neural Networks (CNN)~\cite{jiang2015human}, Long-short Term Memory (LSTM)~\cite{guan2017ensembles} and others~\cite{wang2019deep}.

\begin{figure}[htbp]
  \centering
  \includegraphics[width=.85\textwidth]{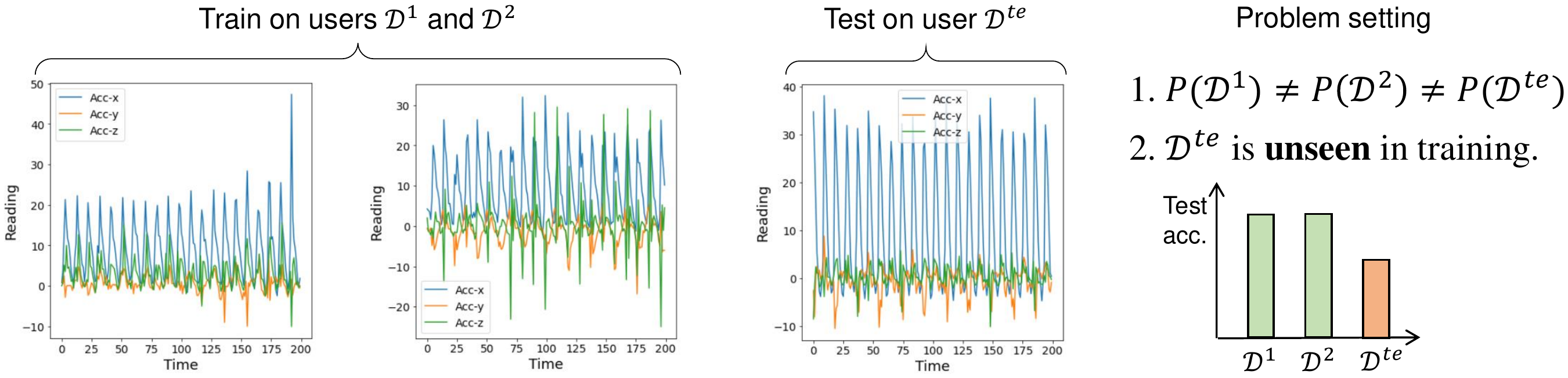}
  \caption{The distributions of accelerometer readings vary among different users. When the test dataset has different distributions and cannot be accessed during training, the performance of existing methods will drop while our method can reach better performance.}
  \label{fig-sensor-reading}
\end{figure}

Despite the great success in the past, one critical challenge is the \emph{generalization} ability of the HAR models, i.e., the performance of applying the trained models to a new, \emph{unseen} dataset.
In real applications, the sensor signals are easily influenced by the diverse personalities of end-users such as acting styles, habits, or different body shapes.
When testing on a new end-user whose activity data are never seen in the training set, the performance of the model is likely to drop.
For instance, \figurename~\ref{fig-sensor-reading} shows the sensor readings of two users from DSADS dataset~\cite{barshan2014recognizing} collected using the same device, where the distributions of sensor readings are different, i.e., $P(\mathcal{D}^1) \ne P(\mathcal{D}^2)$. 
When deployed to an unseen test user $\mathcal{D}^{te}$ whose distribution is different from the training set, the performance of the activity recognition model will be likely to drop.
This is due to the domain shift caused by the non-IID (independently and identically distributed) distributions between training and testing datasets~\cite{soleimani2021cross}.

Such a domain shift issue can be mitigated by transfer learning and domain adaptation techniques~\cite{pan2009survey,cook2013transfer}, which have been applied to HAR over the years.
Transfer learning first pre-trains a model from the source dataset and then fine-tunes it on the new test data.
Domain adaptation performs instance reweighting or feature transformation between the training and testing datasets to learn domain-invariant representations where their distribution gap can be minimized~\cite{wang2018stratified,cook2013transfer,chang2020systematic,khan2018scaling,qin2019cross,sanabria2020unsupervised}.
Unfortunately, both transfer learning and domain adaptation require access to target domain training data, which is often not realistic in real applications as we aim to achieve ``Train once, deploy everywhere''.
In real applications, it is hard to obtain the test data~\cite{wang2021generalizing}.
For example, it is impractical to collect each patient's data for medical healthcare, and to collect data on a variety of fall poses for fall detection in advance.

Domain generalization (DG)~\cite{wang2021generalizing} is an emerging research topic in recent years.
DG focuses on utilizing the knowledge from several different domains to build a model that can generalize well to unknown domains.
Many works have been done and make a good performance in the computer vision field.
Unfortunately, we cannot directly apply existing DG methods to our problem due to the characteristics of wearable-sensor-based activity data.
To the best of our knowledge, few existing works focus on the domain generalization problems in HAR and we refer to this as domain-generalized HAR (DGAR).

In this paper, we propose \textbf{Adaptive Feature Fusion for Activity Recognition (\method)} to improve the generalization ability of HAR models (refer to \figurename~\ref{fig-method}). The key of \method is to learn both domain-specific representation and domain-invariant representation and fuse them dynamically in a unified deep neural network. Specifically, domain-invariant representation learning is to capture the general and transferable knowledge from the training domains, and domain-specific representation learning is to learn the specific characteristics of each domain to preserve the diversity of features to improve generalization ability.
Our key assumption is that although we cannot get access to the test data and the sensor readings from different persons are different, they still share some similarities that can be utilized to learn domain-invariant representations.
Therefore, we can learn transferable knowledge while preserving their diversities for generalization.
Our method can be optimized in an end-to-end neural network.
We show the superiority of \method by experimenting on both public HAR datasets and a real application to the diagnosis of attention deficit hyperactivity disorder (ADHD).
Experiments demonstrate that our method significantly outperforms the comparison methods.

The main contributions of this paper are four-fold:
\begin{enumerate}
\item We propose and study a more practical and challenging problem scenario: domain-generalized activity recognition (DGAR), for robust and generalized activity recognition. We thoroughly analyze the reason for this problem, indicating a new research direction.
\item To solve the DGAR problem, we propose a novel algorithm: Adaptive Feature Fusion for Activity Recognition (\method), to learn both domain-invariant and domain-specific deep representations to enhance the generalization capability of the model to unseen datasets.
\item We evaluate \method on three public HAR datasets. Experiments on cross-person activity recognition demonstrate that the proposed \method can significantly outperform the comparison methods.
\item Finally, we apply our \method algorithm to a real-world ADHD problem where it also achieves the best performance.
\end{enumerate}

\section{Related Work}

\label{sec-related}
\subsection{Human Activity Recognition}
Human activity recognition (HAR)~\cite{ravi2005activity} aims at recognizing the activities of people by training machine learning models on the data collected during performing some specific activities. In HAR, the wearable sensor-based human activity recognition has occupied an important position as it is superior in pervasiveness, computational consumption, and privacy preservation with wearable sensors as interface~\cite{chen2012sensor}. So in this paper, we mainly focus on wearable sensor-based activity recognition. Over these years, many efforts have been done to achieve accurate and robust activity recognition including traditional machine learning methods\cite{lara2012survey} combined with feature extraction, and deep learning methods~\cite{wang2019deep,chen2020deep}. However, the conventional methods usually focus on the i.i.d. data situation, i.e. the training data set and testing data set follow the same distribution, this may result in performance degradation when faced with various tasks in real-life applications.

\subsection{Transfer Learning and Domain Adaptation}
In sensor-based HAR, domain shift is a common and must be solved problem due to the variation of devices, locations, personalities, and so on. During the data collection, any change in these factors may result in distribution divergence. Large distribution divergence results in the performance degradation of the trained model. 
Transfer learning, as a representative method of machine learning, is an effective paradigm to solve the domain shift problems, which makes it possible to reuse existing knowledge. The purpose of transfer learning is to apply the knowledge learned in the existing domains to related but different accessible domains during the training process, to improve the ability to solve new tasks~\cite{pan2009survey,weiss2016survey}.

The commonly used approach of transfer learning is pre-training and fine-tuning, i.e., pre-training on the source dataset to get a pre-trained model, and then fine-tuning this model on the new target dataset.
Such a simple and effective approach has been successfully applied to computer vision~\cite{yosinski2014transferable,shao2014transfer}, natural language processing~\cite{devlin2018bert}, and speech recognition~\cite{jaitly2012application,stoian2020analyzing}.
In order to tackle the domain shift challenge, transfer learning has become an effective method and applied in HAR~\cite{cook2013transfer,chang2020systematic,khan2018scaling,qin2019cross,sanabria2020unsupervised}. Ma M et.al~\cite{ma2016going} proposed a twin stream network architecture and jointly fine-tuned the two networks to recognize objects, actions, and activities. Wang H B et.al~\cite{wang2018domain} verified that fine-tuning and regular constraints can increase the training efficiency, and fine-tuning is valid in practical HAR applications. 

In the scope of transfer learning, domain adaptation~(DA) is a major technique and has attracted much attention from researchers in recent years. Domain adaptation aims at improving the performance on the given less annotated or no annotated target domain by exploiting the source domains. Domain adaptation is a popular topic and has been applied in HAR to solve the domain shift problem~\cite{chang2020systematic}.
Khan et al.~\cite{khan2018scaling} proposed a feature-based model HDCNN, which adapts the source and target features after every convolutional and fully connected layer. Chang et al.~\cite{chang2020systematic} made a comparison of adaptation techniques to give a guideline on applying unsupervised domain adaptation algorithms to cross-position HAR problems.

However, both transfer learning and domain adaptation assume the target domain can be accessed in the model training process. While in real HAR applications, the target tasks are novel and various and can not be accessed during the training. This requires the trained model has strong generalization capability so that it can perform well in unknown fields, which goes beyond the scope of conventional transfer learning and domain adaptation problem settings.
Specifically, the multi-source domain adaptation (MSDA) also has several source domains for training, but its goal is to adapt the trained model to the target domain, which is accessible during training.

\subsection{Domain Generalization}

Domain generalization~(DG) is an emerging topic and is attracting increasing attention in recent years. The goal of DG is to learn a robust and well-generalized prediction function on several given source domains to obtain the minimum prediction error on any possible unknown domain~\cite{wang2021generalizing}. The most striking difference between DA and DG is that whether the target domain can be accessed during the training, the target data can only be used for the model test in DG. Thus, DG is more suitable for the situation of various unknown target HAR tasks in real life. DG has been widely applied in the computer vision field and many DG methods have been proposed and evaluated on image datasets. The DG methods can be mainly divided into three branches, i.e. data manipulation~\cite{shankar2018generalizing,qiao2020learning}, which works on the input data to assist the general representation learning; representation learning~\cite{ghifary2015domain,li2017domain,ganin2016domain}, which aims at learning domain-invariant representation or disentangling the features; and learning strategy~\cite{li2018learning,carlucci2019domain,huang2020self,sagawa2019distributionally}, which exploits the general learning strategy such as meta-learning and ensemble learning for generalization. Yan et al.~\cite{yan2020improve} solved the DG problem from the perspective of data generation by linear interpolation between instances and their labels. However, unlike image data, it is hard to intuitively assess the semantics and diversity of sensor data. Although some works assess sensor data by training a post-hoc classification or prediction model or using other techniques, these assessments are still less intuitive and explainable to some extent~\cite{yoon2019time, li2020activitygan}. Data generation methods may be not straightly applicable to sensor-based HAR since there is still a lack of the intuitive quantitative assessment of quality for generated sensor data. Li et al.~\cite{li2018learning} proposed a model agnostic training procedure by leveraging the meta-learning for DG, i.e. MLDG, while this kind of method may not straightly applicable due to the high dependence of sensor data. Many methods follow distribution alignment in DA to minimize the distribution discrepancy between domains by adversarial training~\cite{gong2019dlow}, Wasserstein distance~\cite{zhou2020domain} etc. to learn domain-invariant representations, and the spirit is followed in DG. In the other aspect, some ensemble-learning-based methods focus on the domain-specific, such as domain-specific neural networks~\cite{wang2020dofe, mancini2018best}, domain-specific batch normalization~\cite{segu2020batch}, weight averaging~\cite{cha2021domain} etc., more details can be found in \cite{wang2021generalizing}. However, few works both consider domain-invariant and domain-specific representations, especially for HAR applications.

Last, it is worth noting that domain generalization is not Leave-One-Out-Cross-Validation (LOOCV) in traditional machine learning. For domain generalization, we sequentially leave one domain for the final test to construct several domain generalization tasks, and these tasks are independent of each other. LOOCV is mainly used for selecting models where all domains are used for validation in turn and it also has an inaccessible test dataset for the final testing.
Therefore, LOOCV can be seen as a typical model selection method that can also be used in DG.

\section{Our Method: Adaptive Feature Fusion}
\label{sec-method}

In this section, we present our Adaptive Feature Fusion method for activity recognition in detail.

\subsection{Problem Definition: Domain-generalized Activity Recognition}

In a typical human activity recognition (HAR) problem, we are given a training dataset $\mathcal{D}^{tr}=\{(\mathbf{x}_i, y_i)\}_{i=1}^{n}$, where $\mathbf{x} \in \mathbb{R}^d$ denotes its $d$-dimensional features and $y \in \{1, 2, \cdots, C\}$ denotes its corresponding activity categories, such as walking or running. $n$ denotes the total number of samples.
The goal is to build a machine learning model $h: \mathbf{x} \mapsto y$ such that it can accurately recognize the activities in the training data, i.e., achieving the minimum training error:
\begin{equation}
    h^\ast = \arg \min_{h} \frac{1}{n} \sum_{i=1}^{n} \ell(h(\mathbf{x}_i), y_i),
\end{equation}
where $h^\ast$ denotes the optimal model and $\ell(\cdot, \cdot)$ is the loss function such as cross-entropy loss.

However, achieving the minimum error on the training data $\mathcal{D}^{tr}$ does not necessarily guarantee optimal performance when we apply the model to the \textbf{unseen} test data $\mathcal{D}^{te}$.
For instance, a well-trained HAR model can perform poorly when deployed to recognize different persons' activities with different body shapes or activity styles.
Moreover, we can never collect all the possible training data to build a generalized HAR model.
While transfer learning and domain adaptation~\cite{qin2019cross,wang2018stratified,pan2009survey,cook2013transfer} are popular to perform cross-domain learning, they can not be used in our problem since they require the availability of the test domain.

We aim to solve this practical and challenging problem, which we refer to as \textbf{Domain-generalized Activity Recognition}, or \textbf{DGAR}.
Here, ``domain'' is a general notion of ``dataset'', i.e., a dataset is a domain, or it can be split into several domains~\cite{wang2021generalizing}.
In DGAR, we assume there are several training (source) domains available, i.e., there are $K$ different but related training domains~$\mathcal{D}^{tr}=\left\{\mathcal{D}^{1}, \mathcal{D}^{2}, \cdots, \mathcal{D}^{K} \right\}$ available.
$\mathcal{D}^{k}=\left\{(\mathbf{x}^{k}_{i}, y^{k}_{i}) \right\}_{i=1}^{n_k}$ denotes the $k_{th}$ training domain with $n_k$ samples.
Our goal is to learn a generalized model $h$ on the $K$ training domains such that it can achieve minimum error on the \emph{unseen} test domain $\mathcal{D}^{te}=\{(\mathbf{x}_i,y_i)\}_{i=1}^{n_{te}}$.
We often assume that all domains share the same kinds of sensors and activities, i.e. the feature space and the label space are the same: $\mathcal{X}^{tr}=\mathcal{X}^{te}$ and $\mathcal{Y}^{tr}=\mathcal{Y}^{te}$. In real applications, different domains tend to have \textit{different} probability distributions, i.e. $P^i(\mathbf{x}) \ne P^j(\mathbf{x}) \ne P^{te}(\mathbf{x}), 1 \le i \le j \le K$.
Since the probability distribution of the test dataset is different from the training domains, DGAR is a practical setting to evaluate the generalization ability of activity recognition algorithms.

\subsection{Motivation and Main Idea}
Domain generalization (DG)~\cite{blanchard2011generalizing, wang2021generalizing} is the general learning setting of DGAR.
Over the past few years, domain generalization has attracted the increasing attention of researchers, and a large number of DG methods have been proposed~\cite{wang2021generalizing}.
Most of these methods are developed for general-purpose learning and they are often evaluated on image classification tasks, with data augmentation~\cite{carlucci2019domain}, domain-invariant representation learning~\cite{matsuura2020domain}, or meta-learning methods~\cite{li2018learning}.
Intuitively, it is natural to ask: \emph{can we directly apply these existing DG methods to our DGAR problem?} The short answer is yes but no. ``Yes'' means we can always do that, ``no'' means this ignores the characteristics of activity recognition and we can do it better.
The reasons are as follows.

First, to achieve strong generalization capability, some of them focus on manipulation of data including data augmentation and data generation which are specific to the image domain~\cite{carlucci2019domain}. However, such methods may not be straightly applicable to sensor-based HAR due to the lack of a quantitative assessment of quality for generated sensor data.
Second, some focus on representation learning, which aims to learn a better feature representation for better generalization. However, activity recognition is a special area where we should not only care about domain-invariant features but also domain-specific features to capture the individually-specific features from each domain, which can preserve the diversity of different persons.
Third, although meta-learning-based DG methods could be used for our problem, our empirical experiments (ref. \tablename~\ref{tab-acc-ds}) indicate that its performance is even worse than the empirical risk minimization method. The reason may be the second-order gradient optimization can have mode collapse for the special activity data.
To sum up, we need to develop special algorithms for this DGAR problem.

In this paper, we proposed a novel \textbf{Adaptive Feature Fusion} method for domain-generalized Activity Recognition, abbreviated as \textbf{\method}.
Our key assumption is that although we cannot get access to the test data in training, we can still learn to represent the test data using the aggregation of existing training data.
This is reasonable since different persons may generate different sensor readings while performing the same activities, but they could share some similarities such as body shapes and activity styles.
Thus, we can learn to represent each data as the weighted aggregation of existing training domains.
Meanwhile, since the activity data from different domains have different probability distributions, we also need to learn domain-invariant features to regularize the model to facilitate knowledge transfer.

\begin{figure}[htbp]
  \centering
  \includegraphics[width=\textwidth]{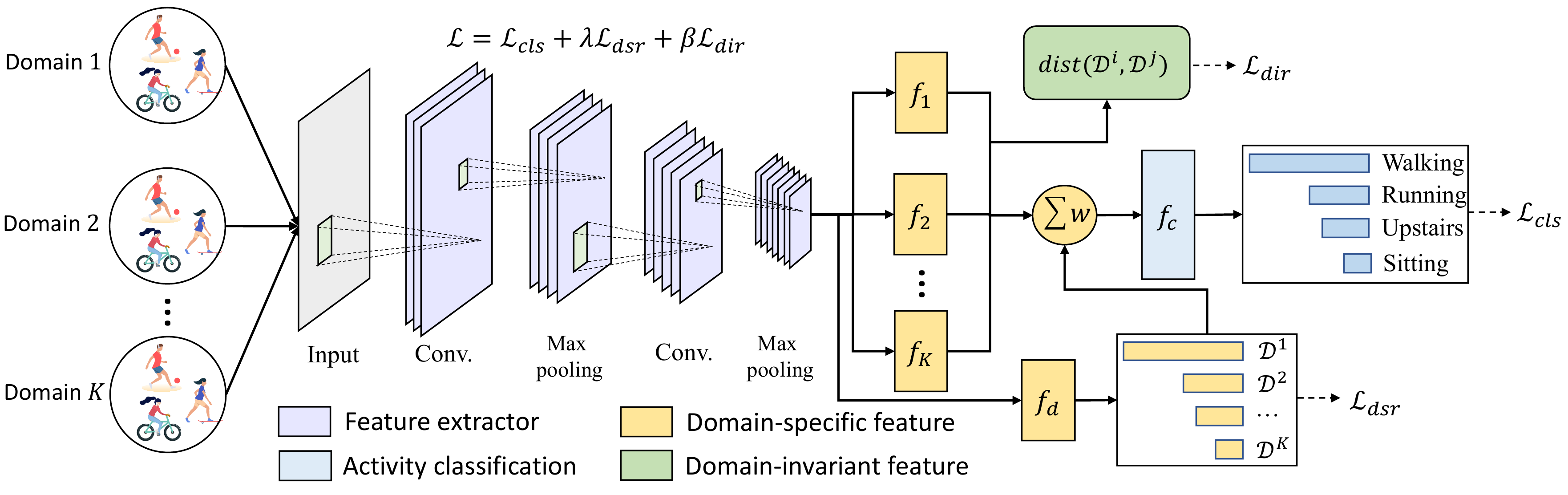}
  \caption{Illustration of the \method framework.}
  \label{fig-method}
\end{figure}

The core of \method is to learn both domain-invariant and domain-specific feature representations. Specifically, domain-invariant representation learning is to capture the general and transferable knowledge from the training domains, and domain-specific representation learning is to learn the specific characteristics of each source domain to enhance the generalization ability. We depict the overall learning process of \method in \figurename~\ref{fig-method}.

\method conceptually consists of four modules: feature extraction module (purple blocks), activity classification module (blue blocks), domain-specific representation learning module (yellow blocks), and domain-invariant learning module (green blocks).
The feature extraction module is used to extract features from the raw sensor data.
In this paper, we leverage two convolutional neural network (CNN) layers along with max-pooling operations to extract features.
These layers are shared by all training domains to reduce parameter amounts.
The domain-specific representation learning module is used to learn domain-specific features, thus this is not shared, but specific for each domain.
We implement it by adding $K$ fully connected (FC) layers for each domain after feature extraction.
To aggregate the specific information of each domain, we design a weighting function. 
The domain-invariant representation learning module is used to reduce the distribution discrepancy between each domain $\mathcal{D}^i$ and domain $\mathcal{D}^j$ to learn domain-invariant features.
Finally, the activity classification module is an FC layer.
Since the classification uses the fusion of domain-specific and domain-invariant features, we call our method adaptive feature fusion.

The learning objective of \method can be formulated as:
\begin{equation}
    \label{eq-all}
    \mathcal{L} = \mathcal{L}_{cls} + \lambda\mathcal{L}_{dsr} + \beta\mathcal{L}_{dir},
\end{equation}
where $\mathcal{L}_{cls}$ is the classification loss, $\mathcal{L}_{dsr}$ is the loss of domain-specific representation learning, $\mathcal{L}_{dir}$ is the loss of domain-invariant representation learning. $\lambda$ and $\beta$ are the tradeoff hyperparameters. For classification, we take cross entropy as the classification loss:
\begin{equation}
    \label{equ-clcloss}
    \mathcal{L}_{cls}=-\frac{1}{N} \sum_{i=1}^{N} y_{i} \log P\left(y_{i} \mid x_{i}\right),
\end{equation}
where $N=\sum_{k=1}^{K} n_k$ is the amount of the samples from all the training domains.

In the next sections, we will elaborate on the details of domain-specific and domain-invariant representation learning modules.

\subsection{Domain-specific Representation Learning}
The domain-specific representation learning module aims to learn domain-specific features and then aggregate them for unified feature representations by fusing the features from multiple sources for the unseen target feature.
More formally, given a new test data $\mathbf{x}$, its feature $\mathbf{z}$ is formulated as:
\begin{equation}
\label{eq-weight-feature}
    \mathbf{z} = \sum_{k=1}^{K} w_k f_k(f_e(\mathbf{x})), \text{~where~}w_k > 0 \text{~and~} \sum_{k=1}^K w_k=1,
\end{equation}
where $w_k$ is the weight on domain $\mathcal{D}^k$, indicating the similarity between the data on domain $\mathcal{D}^k$ and the true data. $f_k$ is the feature learning function of domain $\mathcal{D}^k$ and $f_e$ is the shared feature extraction function, i.e., CNNs.

This process can be viewed as a certain type of ensemble learning~\cite{dietterich2002ensemble}, where each specific base learner is trained to ensemble a stable model which can perform well in all aspects.
In our problem, the base learner is composed of two parts: the shared CNN feature extractor $f_e$ and the specific feature extraction layer $f_k$ for each domain $\mathcal{D}^k$.
This split is motivated by existing research on the transfer learning ability of deep networks~\cite{yosinski2014transferable} that the lower layers tend to learn low-level and general features while the higher layers tend to learn domain-specific features.

To learn the weights $w_k$ for each domain $\mathcal{D}^k$, we build a domain classifier $f_d: \mathbf{x} \mapsto \mathbb{R}^K$ that takes as input the features after CNN layers, and then use softmax to satisfy Equation~\ref{eq-weight-feature}.
At training time, the domain label $d_k \equiv k$ for each sample is known a priori, thus, the domain-specific loss for each domain $k$ can be computed as:
\begin{equation}
\label{eq-domain-classify}
    \mathcal{L}^k_{dsr} = \frac{1}{n_k} \sum_{i=1}^{n_k} \ell(f_d(f_e(\mathbf{x}_i)), d_k).
\end{equation}

Then, we can get the total domain-specific loss by averaging the losses on all domains as $\mathcal{L}_{dsr} = \frac{1}{K} \sum_{k=1}^K \mathcal{L}_{dsr}^k$.

When testing on the target data, we cannot access the domain label of the unseen target data. After the network extracts the features, the output of the domain weight branch is used as the weight to fuse the target features extracted by each domain-specific branch. It can be understood that the learned domain branch network adaptively fuses the domain-specific features extracted from each source-specific branch to construct the feature representation of the target domain. In this way, \method can adaptively learn feature representation of any unseen domains.

\subsection{Domain-Invariant Representation Learning}

While domain-specific feature learning encourages the model to learn specific information for each domain, the feature distribution gap could also be enlarged due to their diverse representations.
Thus, to enhance the generalization capability, we further design a domain-invariant representation learning module to seek balance with domain-specific representation learning.

Recall that features in the lower level of deep neural networks focus on learning common and low-level features, while the higher layers focus more on specific tasks~\cite{yosinski2014transferable}. Thus, we make feature adaptation to reduce the distribution discrepancy between source domains in the domain-specific layers. In the domain adaptation field, the strategy is to reduce the distribution discrepancy between the source domain and the target domain so that the model can be robust on the target. Under domain generalization scenarios, we cannot access the target domain data during the training process, so it is of vital importance to learn domain-invariant feature representation so that any unseen target can be represented. To enable feature adaptation in DGAR, we turn to reducing the distribution divergence between each domain pair $\mathcal{D}^i$ and $\mathcal{D}^j$, i.e., to minimize $dist(\mathcal{D}^i, \mathcal{D}^j)$ where $dist(\cdot, \cdot)$ is a distribution distance measurement.

Specifically, we adopt the widely used distance metric Maximum Mean Discrepancy~(MMD)~\cite{gretton2012kernel} to help reduce the distribution divergence. MMD embeds distributions in Reproducing Kernel Hilbert Space~(RKHS) and calculates the distance between these embeddings as the test statistic. Thus, it is often adopted to justify whether two distributions are the same or used to measure how similar two distributions are. The MMD loss between domains $\mathcal{D}^i$ and $\mathcal{D}^j$ is formulated as:
\begin{equation}
    \label{equ-mmd}
	\mathcal{L}_{dir}^{ij} = 
	\left \Vert \frac{1}{{n}_i}\sum_{\mathbf{x}\in \mathcal{D}^i}{\phi}~(\mathbf{x}) -\frac{1}{{n}_j}\sum_{\mathbf{x}\in \mathcal{D}^j}{\phi}~(\mathbf{x}) \right \Vert_{\mathcal{H}}^{2},
\end{equation}
where $i$ and $j$ are indexes of domains, $\phi \left( \cdot \right) $ is the feature map which maps the original instances into the RKHS $\mathcal{H}$. Then, the total loss for domain-invariant learning can be computed by taking average on all possible domain pairs $\mathcal{L}_{dir}=\frac{2}{K(K-1)} \sum_{i,j} \mathcal{L}^{ij}_{dir}$.

Although we adopt MMD as a metric for distribution divergence in this paper, \method is a general approach that can embed other metrics instead of MMD.

\subsection{Training and Inference}

As for training, after loading the training data, the feature extractor extracts the lower-level features and then inputs them to each domain-specific branch. Each branch learns the higher-level domain-specific representations. Meanwhile, the domain classifier takes the lower-level features as input and outputs the softmax weights to fuse the domain-specific features. At the same time, it makes the in-between-source domain adaptation to learn domain-invariant representation. Then, it takes the fused features as input of the final activity classifier to make activity classification.

As for inference, we fix the model parameters and learn domain-specific and domain-invariant representations for the target. Different from the training, without prior knowledge of the domain label of target data, the output of the domain classifier can be regarded as the similarity between the target and each source domain-specific branch.

The complete learning process of \method is summarized in Algorithm~\ref{algo-AT}.

\begin{algorithm}[htbp!]
	\caption{\method for domain-generalized activity recognition}
	\label{algo-AT}
	\renewcommand{\algorithmicrequire}{\textbf{Input:}} 
	\renewcommand{\algorithmicensure}{\textbf{Output:}} 
	\begin{algorithmic}[1]
		\REQUIRE
		$K$ training domains $\mathcal{D}^{1}_{s},\cdot \cdot \cdot ,~\mathcal {D}^{K}_{s}$, and $\lambda,\beta$.\\
		\ENSURE
		Classification results on test domain.\\
		\STATE Randomly initialize the model parameters $\theta$;\\
		\WHILE{not converge}
		    \STATE Sample a mini-batch $\mathcal{B}=\{\mathcal{B}^1, \cdots, \mathcal{B}^K\}$ from $K$ domains;\\
		    \STATE Extract the lower-level features $f_e(\mathbf{x})$ by the feature extractor;\\
		    \STATE Extract the domain-specific features $f_k(f_e(\mathbf{x}))$ by $K$ domain-specific FC layers;\\
		    \STATE Calculate the domain-specific loss $\mathcal{L}_{dsr}$ and output the weight for each source branch;\\
		    \STATE Calculate the domain-invariant loss $\mathcal{L}_{dir}$.
            \STATE Fuse the domain-specific features with weight according to Eq.~\eqref{eq-weight-feature};\\
		    \STATE Calculate the total loss of \method according to Eq.~\eqref{eq-all}; 
		    \STATE Update the model parameter $\theta$ using SGD.\\
		\ENDWHILE
		\STATE Make inference on the target HAR data.
		\RETURN Classification results on target HAR data.
	\end{algorithmic}
\end{algorithm}

\subsection{Discussions}
\label{sec-diss}
The proposed \method learns domain-specific and domain-invariant representations in a unified framework to seek their balance in feature learning, which can take other distribution discrepancies as the domain-invariant learning loss.
We show two possible losses: the domain-adversarial neural networks (DANN)~\cite{ganin2016domain} and the COReration ALignment (CORAL) loss~\cite{sun2016deep}.
DANN introduced an adversarial training objective where it used a min-max optimization to maximize the loss of domain discriminator and minimize the loss of both feature extractor and classification.
However, DANN only aims at learning domain-invariant representations, which is good for domain adaptation tasks (that explains why DANN is the base model for modern domain adaptation models). DANN ignores the specific features for each domain that is useful for domain generalization tasks, making it less favorable for our DGAR problem.
We will empirically show this argument in later experiments (ref. Section~\ref{sec-extend-dir}).
On the other hand, \method is a general framework for DGAR tasks where we can also employ the domain discriminator loss of DANN to replace MMD:
\begin{equation}
\label{eq-dann}
    \mathcal{L}^{dann}_{dir} = \mathbb{E}_{1 \le i \ne j \le K} \mathbb{E}_{\mathbf{x}^i \in \mathcal{D}^i, \mathbf{x}^j \in \mathcal{D}^j}\log [D(f_i(f_e(\mathbf{x}^i)))] + \log [1- D(f_j(f_e(\mathbf{x}^j)))],
\end{equation}
where $\mathbb{E}$ denotes expectation operation and $D$ is a domain discriminator (typically a two-layer feed-forward network). Thus, DANN for domain generalization requires to build $\frac{K(K-1)}{2}$ domain discriminators, which is not efficient.
We can also replace MMD with CORAL loss as:
\begin{equation}
    \label{eq-coral}
    \mathcal{L}^{coral}_{dir} = \frac{1}{4d^2}||C^i-C^j||^2_F,
\end{equation}
where $d$ denotes the feature dimension, $C^i, C^j$ denotes the covariance matrices for two domains, and $||\cdot||_F$ denotes the Frobenius norm.
In later experiments (Section~\ref{sec-extend-dir}), we will show that our \method can also achieve competitive performance with these two losses.

\subsection{Theoretical Insights}

Finally, we show that our algorithm is theoretically-motivated using the theory proposed in \cite{albuquerque2019generalizing}.

\begin{theorem}[Risk upper bound on unseen domain~\cite{albuquerque2019generalizing}]
Let $\gamma=d_\mathcal{H}(\mathcal{D}^{te}, \bar{\mathcal{D}}^{te})$ denote the $\mathcal{H}$-divergence between target domain and its nearest neighbor in source domain convex hull, then, the risk on unseen domain $\mathcal{D}^{te}$ of hypothesis $h$ is upper-bounded by the weighted risk on source set $S$:
\begin{equation}
    R_{te}[h] \leq \sum_{i=1}^{N_{S}} \pi_{i} R_{S}^{i}[h]+\gamma+\epsilon+\min \left\{\mathbb{E}_{\bar{\mathcal{D}}^{te}}\left[\left|f_{S_{\pi}}-f^{te}\right|\right], \mathbb{E}_{\mathcal{D}^{te}}\left[\left|f^{te}-f_{S_{\pi}}\right|\right]\right\},
\end{equation}
where $\epsilon$ is the largest distribution divergence between unseen target domain and any source domain and $\min \left\{\mathbb{E}_{\bar{\mathcal{D}}^{te}}\left[\left|f_{S_{\pi}}-f^{te}\right|\right], \mathbb{E}_{\mathcal{D}^{te}}\left[\left|f^{te}-f_{S_{\pi}}\right|\right]\right\}$ denotes the difference between labeling functions.
\end{theorem}

In our problem, the categories between training and testing are the same, the main distribution difference between training and testing data is the activity patterns (i.e., $P(\mathbf{x})$). So it is close to the covariate shift assumption: the labeling function error ($\min \{\cdot, \cdot\}$) and $\gamma$ are both relatively small \cite{albuquerque2019generalizing}.
In this way, the risk on unseen domain is bounded by two terms: the weighted source risk $\sum_{i=1}^{N_{S}} \pi_{i} R_{S}^{i}[h]$ and the source-target distribution divergence $\epsilon$. Obviously, our domain-specific learning module (Eq.~\eqref{eq-weight-feature}) corresponds to minimizing the weighted source risk and the domain-invariant learning module (Eq.~\eqref{equ-mmd}) minimizes the risk $\epsilon$. Thus, our algorithm can also be interpreted from the theory. Additionally, we also provide a visualization study to help better analyze the algorithm in Section~\ref{sec-exp-ablation}.

\section{Experimental Evaluation}
\label{sec-exp}
In this section, we evaluate the performance of the proposed \method approach via extensive experiments on domain-generalized activity recognition.

\subsection{Datasets and Preprocessing}
\label{subsec-preprocess}
We adopt three large public activity datasets as summarized in \tablename~\ref{tb-dataset}. In the following, we briefly introduce their basic information, and detailed descriptions are in their original papers.

\begin{table}[htbp]
\vspace{-.1in}
  \caption{Statistical information of three public activity recognition datasets}
  \vspace{-.1in}
  \label{tb-dataset}
  \resizebox{.9\textwidth}{!}{
  \begin{tabular}{cccccccc}
    \toprule
    Dataset&Subject&Activity&Sample&Body Position&Sampling rate & Training : Test size\\
    \midrule
            DSADS  &8  &19  &1.14M  &5  &25Hz & $\sim$15:1\\ 
			USC-HAD  &14  &12  &2.81M  &1  &$\sim$100Hz & $\sim$16:1\\ 
			PAMAP2  &9  &18  &2.84M &3  &100Hz & $\sim$15:1\\ 

  \bottomrule
\end{tabular}}
\end{table}

\paragraph{DSADS}
UCI Daily and Sports Data Set~\cite{barshan2014recognizing} collects 19 activities through 8 subjects (four males and four females between the ages of 20 and 30) wearing body-worn sensor units including triaxial accelerometer, triaxial gyroscope, and triaxial magnetometer on 5 body parts: torso, right arm, left arm, right leg and left leg. To construct the domain-generalized activity recognition scenario, we divide the 8 subjects into 4 groups where each group consists of 2 different subjects, and data from each group is regarded as a domain, leading to 4 domains in total.

\paragraph{USC-HAD}
USC Human Activity Dataset~\cite{zhang2012usc} consists of data collected from 14 subjects (7 males and 7 females) performing 12 activities. A motion mode is equipped at the front right hip of subjects to capture triaxial accelerometer and triaxial gyroscope sensor readings. In order to construct a domain-generalized activity recognition scenario, we divide the 14 subjects into 5 groups where each of the first four groups consists of 3 different subjects, and the last group consists of two subjects. That leads to 5 domains in total.

\paragraph{PAMAP2}
PAMAP2~\cite{reiss2012introducing} consists of data collected from 9 subjects performing 18 activities. Each subject wears 3 inertial measurement units (IMU) and a heart rate monitor. We use the data from IMU in the experiments. Each IMU consists of two 3-axis accelerometers, one 3-axis magnetometer, and one 3-axis gyroscope. To construct a domain-generalized activity scenario, we choose 8 subjects~(subjects IDS 1-8) and their common eight activities: lying, sitting, standing, walking, ascending stairs, descending stairs, vacuum cleaning, and ironing. Data are divided into 4 domains.

For each task in one dataset, we select one domain as the test domain while other domains serve as the training domains.
We further split a validation set from the training set with a ratio of $0.2$ for hyperparameter tuning.
Our main focus is to test the performance on cross-person settings in a dataset (i.e., different person, same sensor device).

\subsection{Comparison Methods and Implementation Details}
We compare \method with several state-of-the-art domain generalization methods:
\begin{itemize}
\item Empirical Risk Minimization~(ERM~\cite{vapnik1992principles}, i.e., CNN baseline): minimizes the sum of errors over data. We regard it as the naive baseline to learn a single model on all source domains.
\item Meta-Learning Domain Generalization~(MLDG~\cite{li2018learning}):
learns how to generalize cross domains by leveraging Model-Agnostic Meta Learning~(MAML~\cite{finn2017model}).
\item Domain Adversarial Neural Network~(DANN~\cite{ganin2016domain}): employs an adversarial network that consists of a generator and a discriminator to adapt feature distribution. Under the domain generalization setting, we perform DANN across multiple source domains as no target can be accessed during the training process. 
\item Group Distributionally Robust Optimization~(GroupDRO~\cite{sagawa2019distributionally}): couples group DRO models with increased regularization to increase the importance of the worst-group loss.
\item Representation Self-Challenging~(RSC~\cite{huang2020self}): discards the dominant features i.e. representations associated with the higher gradients at each epoch, and forces the model to predict with the remaining information.

\item AND-mask~\cite{parascandolo2020learning}: learning explanations that are hard to vary, which uses AND-mask to improve the consistency in gradients for better generalization.
\end{itemize}
In addition, we compare the results of all methods with the results trained on the target domain (split the target data into train and test set with a rate around 8:2), which are \emph{ideal} cases since our problem does not access the target domain data:
\begin{itemize}
    \item ERM-t: directly trains models on the target domain using ERM.
    \item Fine-tune: trains a model using ERM on the training set and then fine-tunes it on the target.
\end{itemize}

Experimental settings are as follows. In order to evaluate the classification performance of \method, we construct non-iid cross-person HAR tasks under the domain generalization scenario.
First, we divide the data of subjects into several groups as illustrated in Section~\ref{subsec-preprocess}, data from each group is regarded as a domain. Each domain plays the role of the unseen target domain and remains as source domains.
We use 2-D convolutions for our implementations with the kernel size of $(1, 6)$ and $(1, 9)$, depending on different datasets.

For the comparison methods, we adopt the implementations from DomainBed~\cite{gulrajani2020search} while we change their network structures to be the same as ours for the comparison study.
We perform hyperparameter tuning for each comparison method to achieve its best performance on each task.
Specifically, we tune the following hyperparameters: learning rate is selected in $\{0.0003, .., 0.001\}$, batch size is set as 128.
We set the total training epochs as $500$ and early stop patience to $30$.
We run the experiments five times and report the average results.
F1 score is the main evaluation metric, and we also analyze the accuracy, precision, recall, and ROC curves in detailed analysis.

\begin{table}[b!]
\centering
\caption{Weighted F1 score (\%) on DSADS dataset. The bold is the best result except for two ideal conditions.}
\label{tab-acc-ds}
\resizebox{\textwidth}{!}{
\begin{tabular}{c|ccccccc|cc}
\toprule
Target  & ERM & MLDG & DANN  & GroupDRO  & RSC & AND-mask & \method~(ours) & ERM-t & Fine-tune  \\ \hline
T-1     & 82.58 & 64.19 & \textbf{83.94} & 81.58 & 81.16 & 81.86  & 82.74 & 98.90 & 99.12\\ 
T-2     & 80.04 & 72.32 & 79.52 & 80.75 & 79.86  & 80.78  & \textbf{84.92} & 98.90 & 98.90\\ 
T-3     & 82.87 & 80.79 & 83.36 & 83.24 & 83.53 & 83.24  &\textbf{87.60} & 99.34 & 100 \\ 
T-4     & 82.82 & 52.81 & 84.76 & 82.69 & 84.03 & 80.83  &\textbf{86.45} & 98.22 & 98.22\\ \hline
Average & 82.10 & 67.53 & 82.90 & 82.07 & 82.15 & 81.68 & \textbf{85.43} & 98.84 & 99.06\\ \bottomrule
\end{tabular}}
\end{table}

\begin{table}[t!]
\centering
\caption{Weighted F1 score (\%) on USC-HAD. The bold is the best result except for two ideal conditions.}
\label{tab-acc-uschad}
\resizebox{.9\textwidth}{!}{
\begin{tabular}{c|cccccc|cc}
\toprule
Target  & ERM & DANN  & GroupDRO  & RSC & AND-mask  & \method~(ours) & ERM-t & Fine-tune \\ \hline
T-1     & 72.79 & 74.68 & 73.32 & 73.75 & 72.35  & \textbf{75.11} & 90.80 & 91.22 \\ 
T-2     & 76.87 & 77.64 & 75.82 & 77.12 & 75.17  & \textbf{78.57} & 89.90 & 90.99 \\ 
T-3     & 72.82 & 72.00 & 69.27 & 73.70 & 70.86  & \textbf{74.59} & 87.73 & 89.74 \\ 
T-4     & 59.21 & 60.05 & 57.67 & 59.66 & 58.88  & \textbf{62.20} & 82.67 & 84.72 \\ 
T-5     & 65.83 & 68.18 & 60.68 & 70.17 & 67.38  & \textbf{72.41} & 87.72 & 90.33 \\ \hline
Average & 69.50 & 70.51 & 67.35 & 70.88 & 68.93  & \textbf{72.58} & 87.76 & 89.40 \\ \bottomrule
\end{tabular}}
\end{table}

\begin{table}[t!]
\centering
\caption{Weighted F1 score (\%) on PAMAP2. The bold is the best result except for two ideal conditions.}
\label{tab-acc-pamap}
\resizebox{.9\textwidth}{!}{
\begin{tabular}{c|cccccc|cc}
\toprule
Target  & ERM & DANN  & GroupDRO  & RSC & AND-mask  & \method~(ours) & ERM-t & Fine-tune  \\ \hline
T-1     & 56.25 & 57.92 & 57.55 & 57.63 & 56.74 & \textbf{65.37} & 94.88 & 95.58 \\ 
T-2     & 84.06 & 86.11 & 86.91 & 86.58 & 86.54 & \textbf{89.35} & 93.83 & 94.95 \\ 
T-3     & 85.21 & 86.21 & 85.24 & 84.85 & 84.93 & \textbf{87.43} & 94.04 & 94.70 \\ 
T-4     & 82.72 & 83.32 & 84.02 & 84.12 & 82.28 & \textbf{86.46} & 93.90 & 95.21 \\ \hline
Average & 77.06 & 78.39 & 78.43 & 78.29 & 77.62 & \textbf{82.15} & 94.16 & 95.11 \\ \bottomrule
\end{tabular}
}
\end{table}

\subsection{Classification Performance}

The test weighted F1 score on three datasets are shown in \tablename~\ref{tab-acc-ds}, \ref{tab-acc-uschad}, and \ref{tab-acc-pamap}, respectively.
From these results, we can make some observations that: 1) The proposed \method can achieve the best classification on almost all tasks. Concretely speaking, \method significantly outperforms the second-best comparison methods by 2.5\%, 1.7\%, and 3.7\% on three datasets, respectively. 2) Other comparison methods such as AND-mask and RSC can achieve good classification on some tasks while behaving less satisfied on others, this may be because they neglect the domain-specific knowledge, which may neglect some latent information between the distributions. 3) ERM is regarded as the naive baseline and the experimental results are worse than other methods on several tasks, it is because it only minimizes the empirical risk on the training source data without reducing the distribution discrepancy and investigating the latent information, the generalization capability is less satisfied due to the large distribution discrepancy. Thus, it is necessary to explore and utilize domain generalization approaches in real-life cross-domain HAR application tasks. 4) The performance of DANN is also not comparable to ours. It aims to learn domain-invariant representations. However, it does not consider the domain-specific information, making it less effective for generalization tasks.
5) The average performance of MLDG in \tablename~\ref{tab-acc-ds} is significantly worse than ERM, indicating it is not feasible to directly apply meta-learning-based DG algorithms to activity recognition problems. This may be because the split of meta-train and meta-test datasets depends heavily on the independence of data, while the activity sensor data are highly dependent, thus making the results worse. The same conclusions go for data augmentation methods whose results are also not comparable, thus we did not list them.
6) Finally, the results of the two ideal-case methods ERM-t and Fine-tune which are trained on the target domain are better than all DG methods, indicating the importance of target train data. While \method achieves the best results, there is still room for improvement.

\begin{figure}[t!]
    \centering
    \vspace{-.1in}
    \subfigure[Ablation study]{
    \includegraphics[width=.31\textwidth]{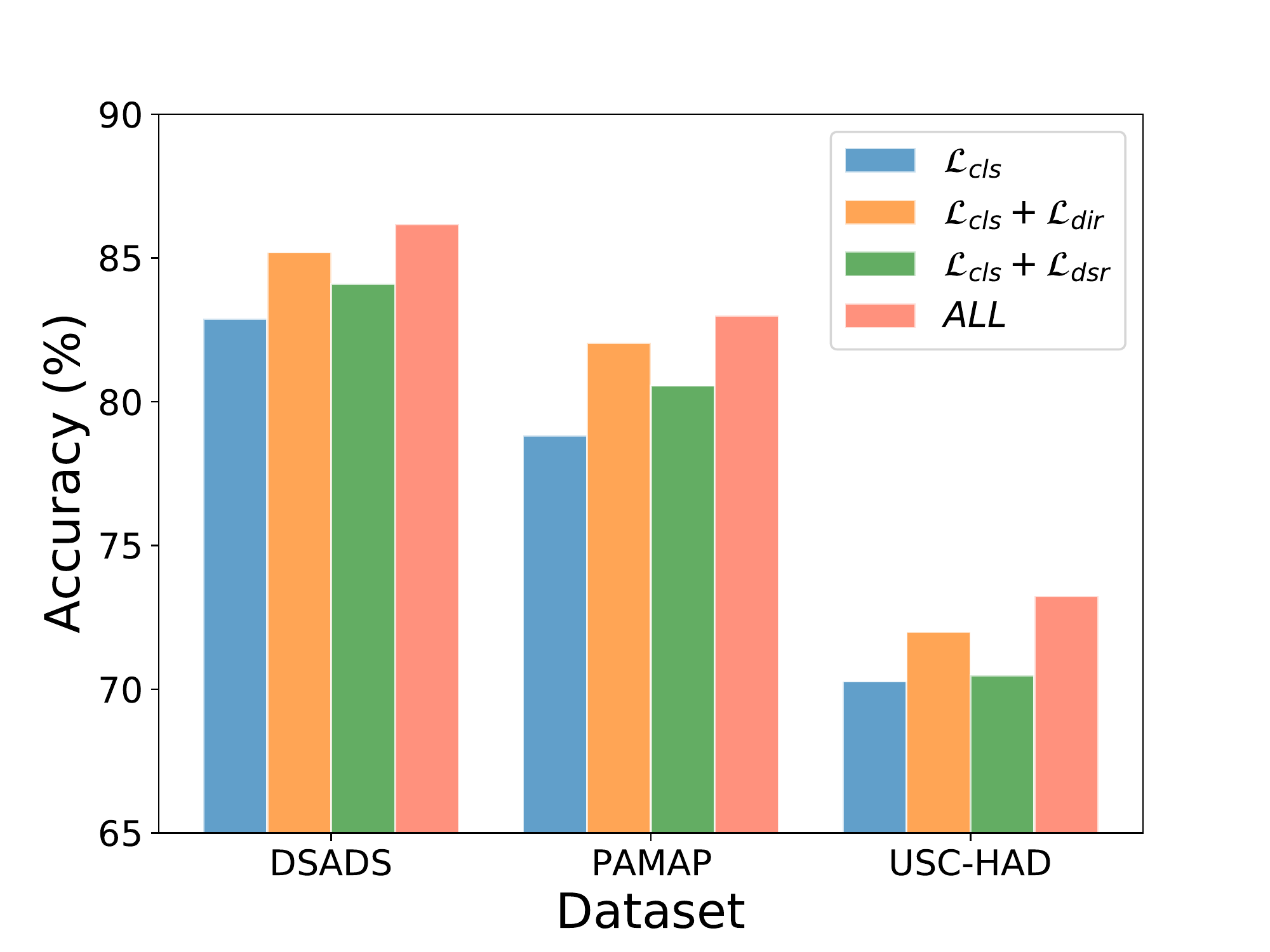}
    \label{fig-ablation-dirdsr}
    }
    \subfigure[\method with DANN and CORAL]{
    \includegraphics[width=.31\textwidth]{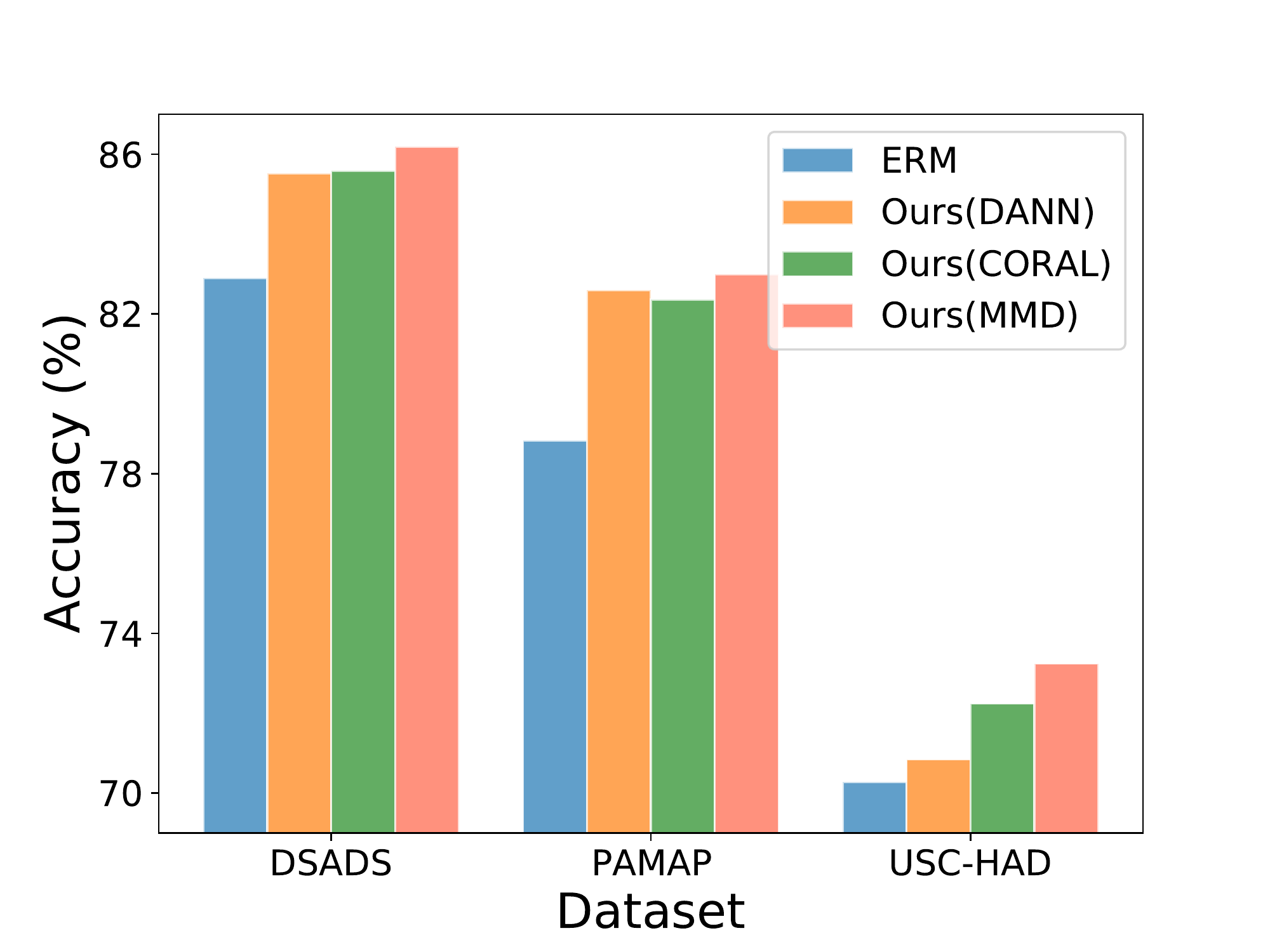}
    \label{fig-ablation-adv}
    }
    \subfigure[Learned weights]{
    \includegraphics[width=.31\textwidth]{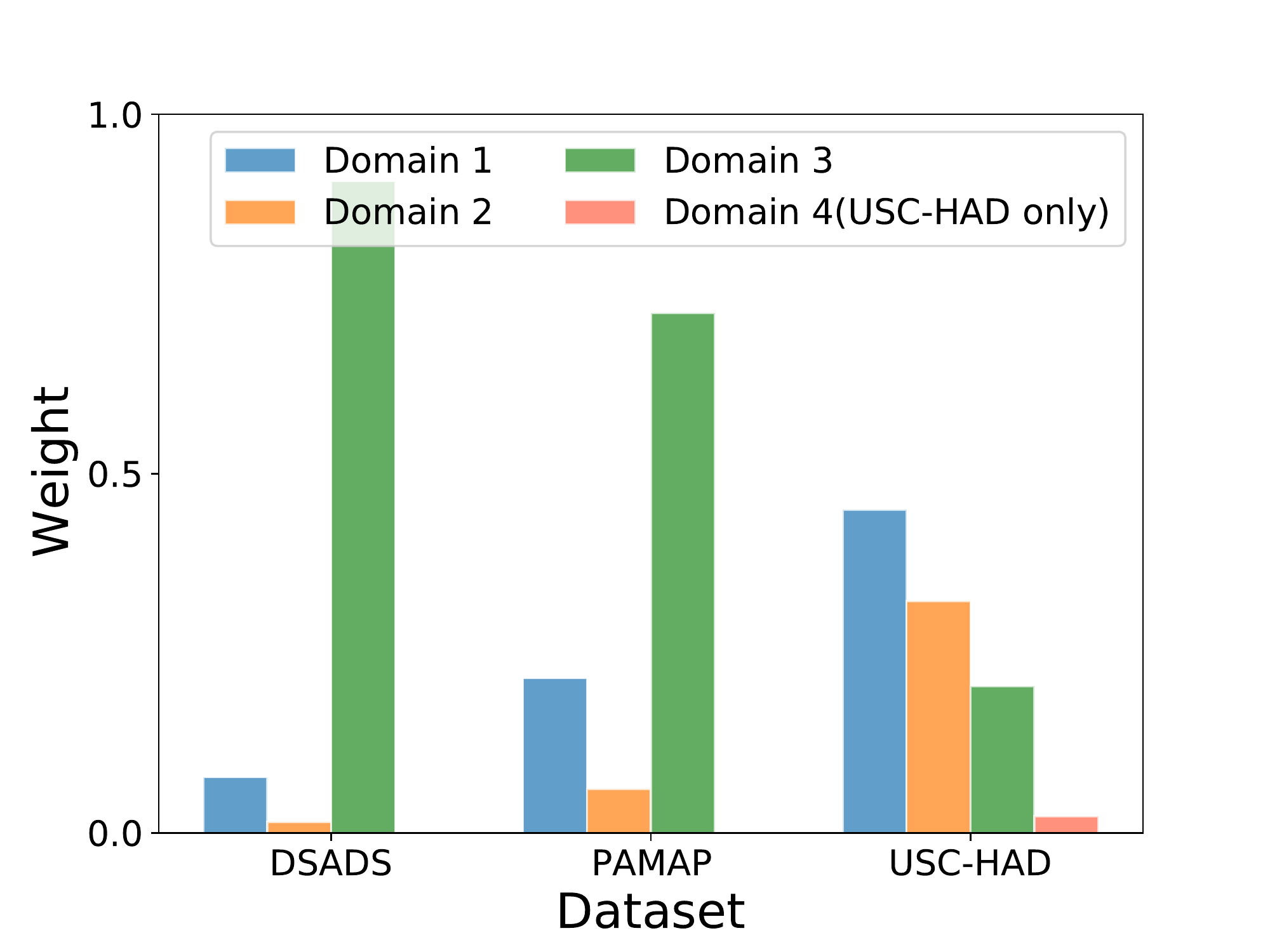}
    \label{fig-weight}
    }
    \caption{Detailed analysis of \method. (a) Ablation study to show the effectiveness of domain-invariant and domain-specific representation learning modules. (b) Replace the MMD loss with DANN and CORAL loss. (c) Weights of a randomly selected target test sample for each dataset.}
    \label{fig-ablation}
\end{figure}

\begin{figure*}[t!]
	\centering
	\vspace{-.2in}
	\subfigure[Only $\mathcal{L}_{cls}$]{
		\includegraphics[scale=0.4]{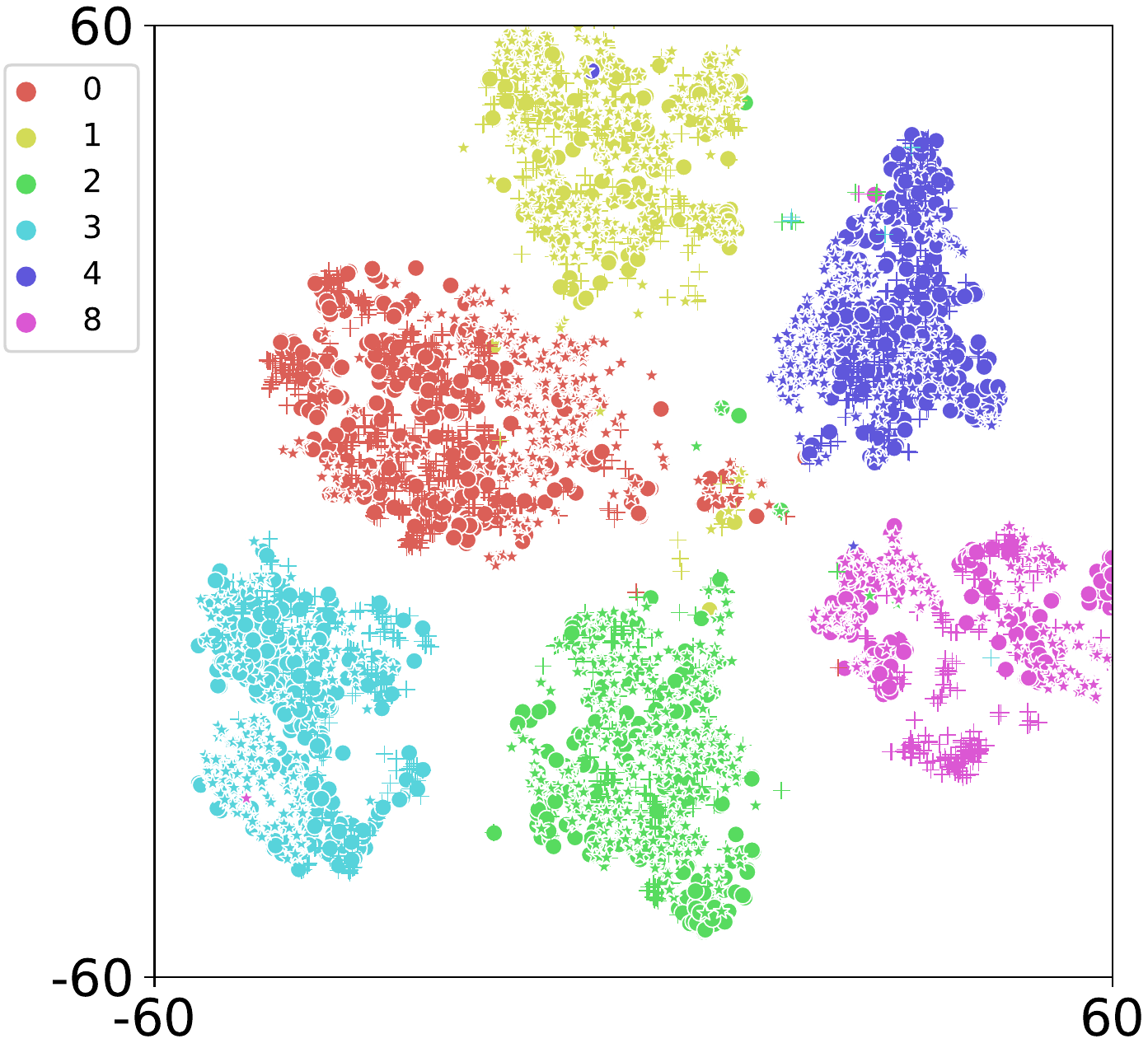}
		\label{fig-tsne-cls}}
	\subfigure[Adding domain-specific learning]{
		\includegraphics[scale=0.4]{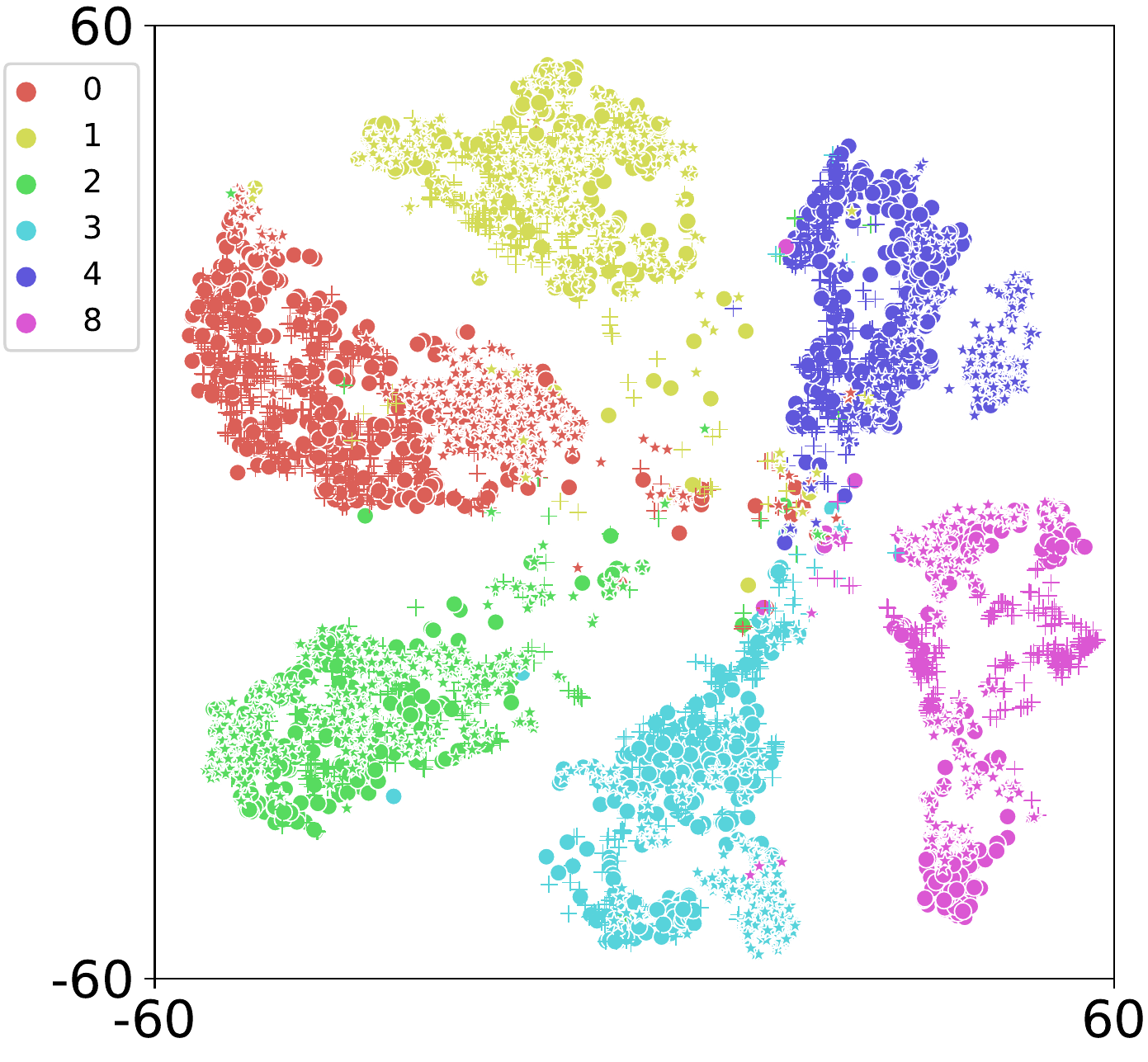}
		\label{fig-tsne-dsr}}
	\vspace{-.2in}
	\subfigure[Adding domain-invariant learning]{
		\includegraphics[scale=0.4]{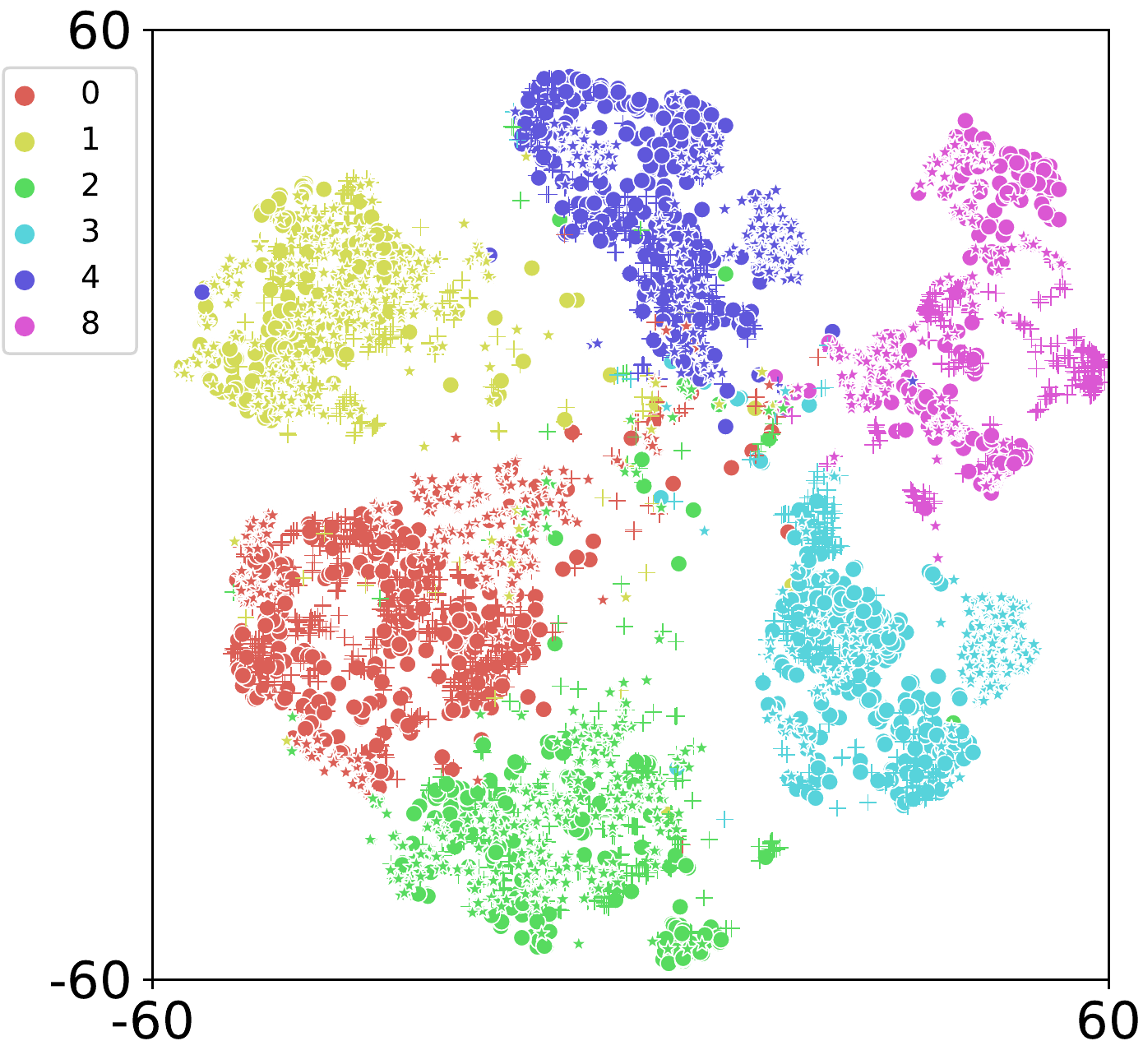}
		\label{fig-tsne-dir}}
	\subfigure[Domain-specific + domain-invariant learning]{
		\includegraphics[scale=0.4]{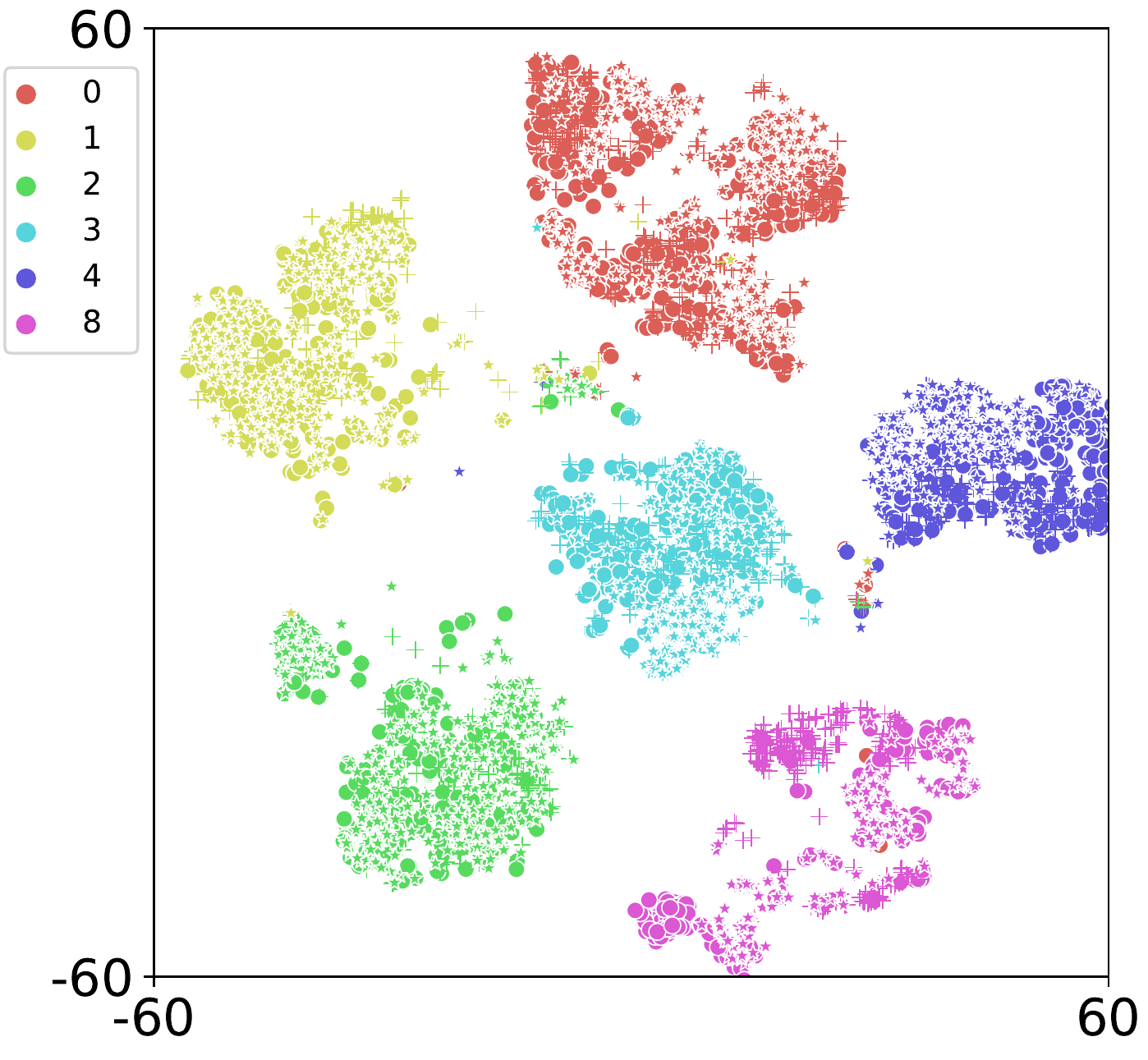}
		\label{fig-tsne-all}}
	\caption{Visualization of the t-SNE embeddings of USC-HAD dataset. Each class is denoted by color and each domain is denoted by a shape. The classes denoted by numbers are walking forward, walking left, walking right, walking upstairs, walking downstairs, and standing. \textsl{Best viewed in color and zoom in.}}
	\label{fig-tsne}
\end{figure*}

\subsection{Ablation Study}
\label{sec-exp-ablation}
\subsubsection{Domain-invariant and domain-specific learning modules}
\method consists of two important modules: (1) domain-invariant learning module and (2) domain-specific learning module. In this section, we conduct an ablation study by evaluating the importance of each module.
We compare four variants of our method: (1) $\mathcal{L}_{cls}$, (2) $\mathcal{L}_{cls} + \mathcal{L}_{dir}$, (3) $\mathcal{L}_{cls} + \mathcal{L}_{dsr}$, and (4) the full version of \method.
\figurename~\ref{fig-ablation-dirdsr} reports the average classification accuracy of these variants on all tasks.
It can be observed that by combining domain-invariant and domain-specific learning modules, the whole \method can achieve the best performance. It evaluates that both the modules are very important and make contributions to the accurate classification of HAR tasks.

We also show the feature embeddings of each module of our method in Figure~\ref{fig-tsne}. (1) Compare \figurename~\ref{fig-tsne-cls} and \figurename~\ref{fig-tsne-dsr}, we see that adding domain-specific learning to the model will enlarge the domain margin and the classes are far from each other. This is because domain-specific learning focuses on separating classes in each domain. However, the domains are not aligned well (in each class, different domains denoted by shapes are still far). (2) Compare \figurename~\ref{fig-tsne-cls} with \figurename~\ref{fig-tsne-dir}, we see that domains (denoted by shapes) are more invariant in each class since domain-invariant learning focuses on learning general features that can transfer across domains. But the classification is worse than the whole version (class margins are small, making it easy to misclassify samples). (3) Finally, we see from \figurename~\ref{fig-tsne-all} that adding two modules can not only make the domains more invariant but also enhance the classification results.

\subsubsection{Extending domain-invariant learning with other distances}
\label{sec-extend-dir}
In Section~\ref{sec-diss}, we show that our \method can also take other distribution matching techniques such as domain-adversarial learning (DANN, Eq.~\eqref{eq-dann}) and CORAL loss (Eq.~\eqref{eq-coral}).
In this section, we replace the original MMD measure with the DANN and CORAL loss to evaluate the performance of \method.

We thoroughly test the performance in three different datasets and record their average accuracy in \figurename~\ref{fig-ablation-adv}.
It shows that our method is general and flexible that can use other distribution matching metrics to achieve competitive performance, which is better than the original ERM.
We also observe that \method with MMD loss gives the best performance. On the other hand, comparing their computational complexity ($\mathcal{O}_{MMD}\approx \mathcal{O}_{CORAL} < \mathcal{O}_{DANN}$), we use MMD as our main distribution matching loss.

\subsubsection{Analysis of domain-specific module}

In this section, we analyze the domain-specific module by investigating the learned domain weights to the target domain.
The weights can act as the similarity between the training domains and the test data, representing how much information can be transferred from these domains.
\figurename~\ref{fig-weight} shows the (normalized) weights given to a target test sample for each dataset.
Note that only the USC-HAD dataset has four training domains and the other two have three training domains.
The weights reflect the similarity between the target dataset and each training domain.
Thus, it shows that our \method can effectively learn such similarity, which acts as the contribution of each domain to the target dataset for better generalization.

\begin{figure*}[t!]
	\centering
	\vspace{-.2in}
	\subfigure[Target-1]{
		\includegraphics[scale=0.4]{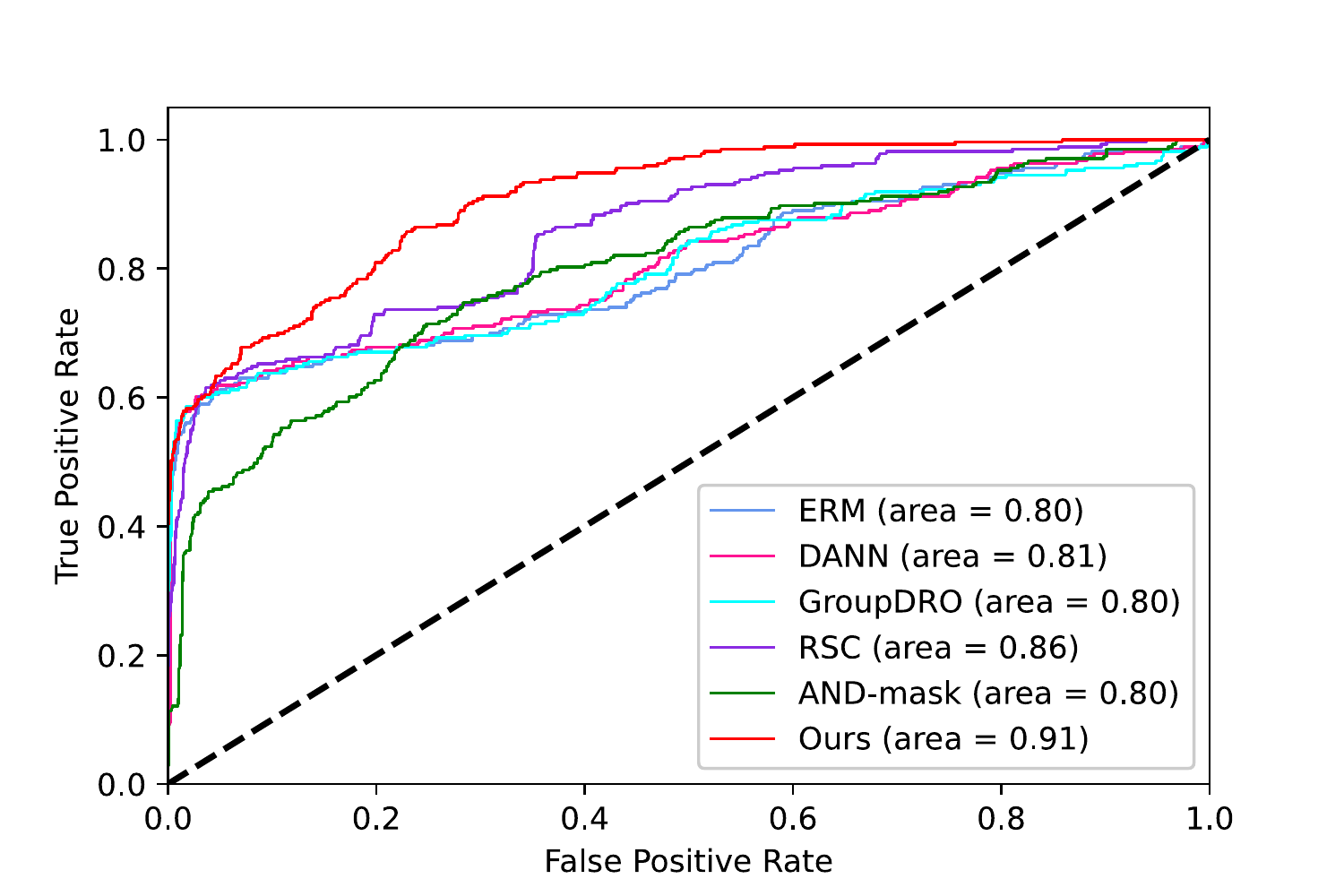}
		\label{fig-roc_pa_t0}}
	\subfigure[Target-2]{
		\includegraphics[scale=0.4]{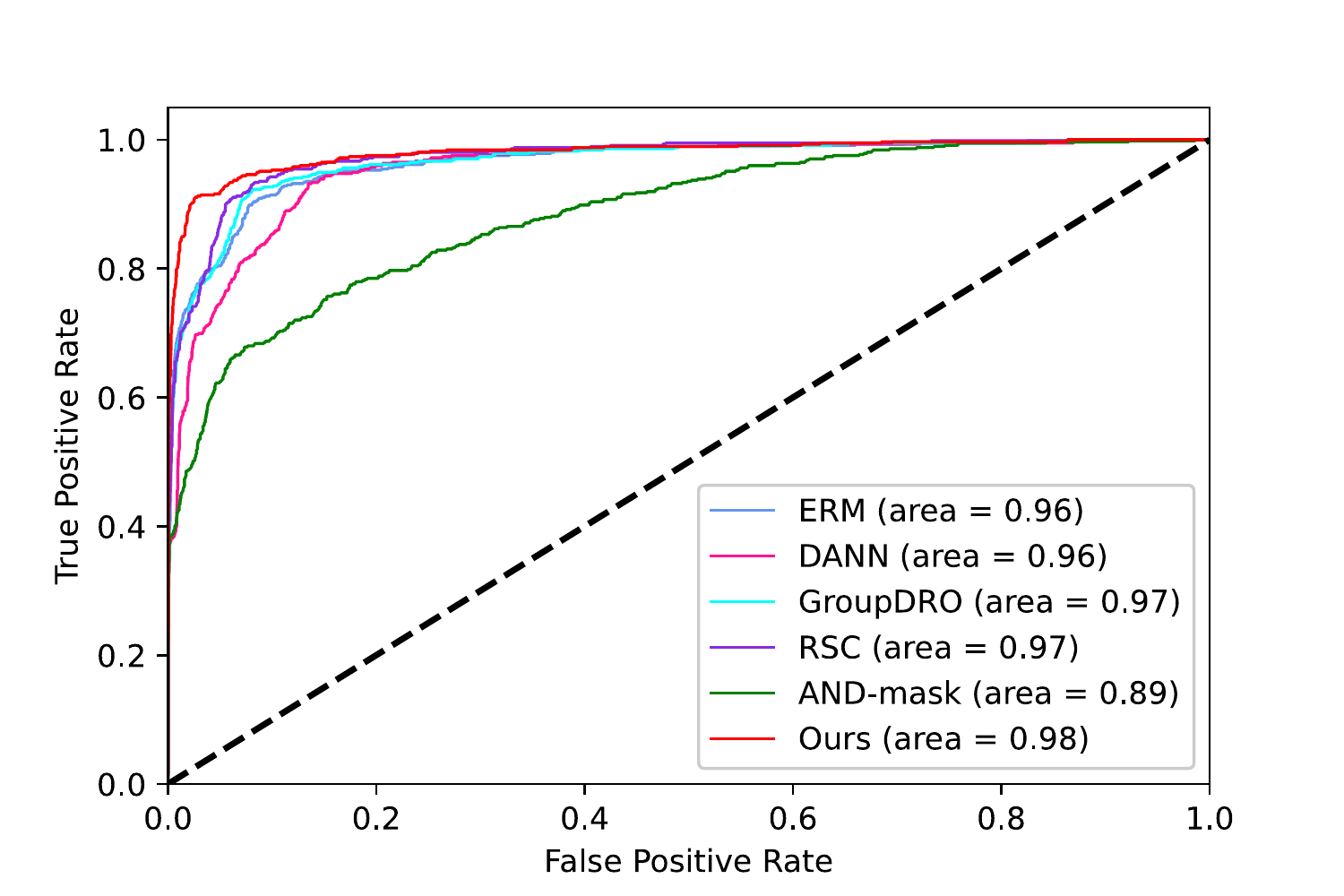}
		\label{fig-roc_pa_t1}}
	\vspace{-.2in}
	\subfigure[Target-3]{
		\includegraphics[scale=0.4]{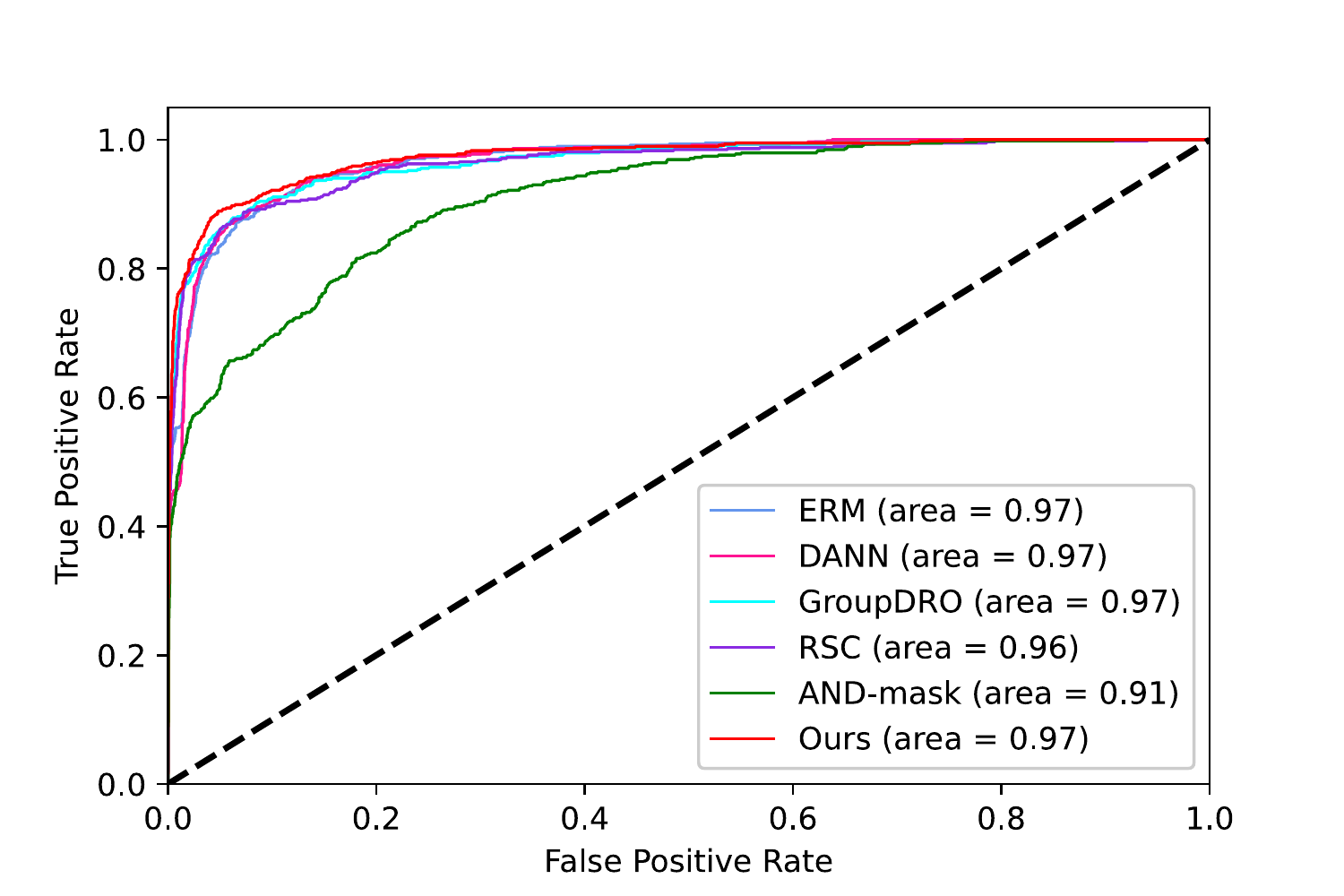}
		\label{fig:roc_pa_t2}}
	\subfigure[Target-4]{
		\includegraphics[scale=0.4]{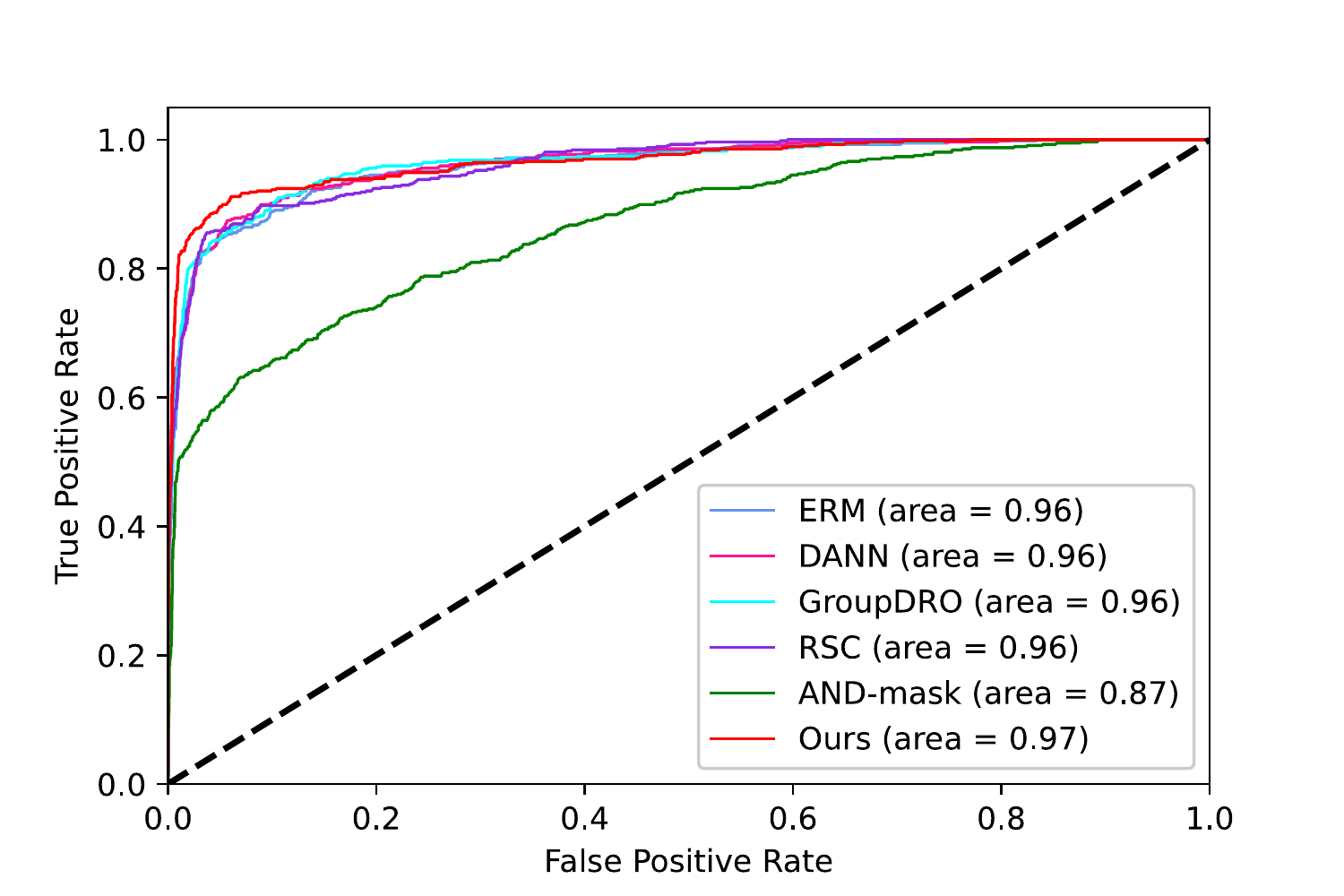}
		\label{fig:roc_pa_t3}}
	\caption{Micro-average ROC curves of \method and other comparison methods for each task on PAMAP2.}
	\vspace{-.1in}
	\label{fig-roc}
\end{figure*}

\subsection{Detailed Analysis}
To further evaluate the performance of \method on each class, we provide the fine-grained analysis by the micro-average Receiver Operating Characteristics~(ROC) curves in \figurename~\ref{fig-roc}.
ROC is a more effective metric for the areas of cost-sensitive learning and unbalanced class issues~\cite{fawcett2006introduction}. We also calculate the AUC (Area Under Curve) on the four tasks of the PAMAP2 dataset. \figurename~\ref{fig-roc} illustrates the following conclusions: 1) Our \method can achieve good performance on four tasks. The micro-average results of four target tasks are all higher than 0.91 which evaluates the classification effectiveness of \method. 2) Results of all methods on Target-1 are less satisfying, which may be because the distribution discrepancy of the other domains is smaller so the generalization capability is reduced. The improvement of \method is more obvious on Target-1 than on the other three target tasks, which shows that our approach is more robust on hard tasks.

\begin{table}[t!]
  \centering
  \caption{Precision, recall, and F1 score of \method and other comparison methods for each class on PAMAP2. Numbers in each cell denote the percentage of the prediction for each class.}
  \vspace{-.1in}
  \label{tb-prf-pamap-t1}
  \resizebox{1\textwidth}{!}{
  \begin{tabular}{c|ccc|ccc|ccc|ccc|ccc}
    \toprule
 & \multicolumn{3}{c}{DANN} & \multicolumn{3}{c}{GroupDRO} & \multicolumn{3}{c}{RSC} & \multicolumn{3}{c}{AND-mask} & \multicolumn{3}{c}{\method} \\
    \midrule
Activity & P & R & F1 &P & R & F1 & P & R & F1 & P & R & F1 & P & R & F1 \\ \hline
Lying & \textbf{1.00} & 0.94 & 0.97 & \textbf{1.00} & \textbf{0.96} & \textbf{0.98} & 0.99 & \textbf{0.96} & 0.97 & 0.99 & \textbf{0.96} & 0.97 & \textbf{1.00} & \textbf{0.96} & \textbf{0.98}   \\
Sitting & 0.83 & \textbf{0.87} & 0.85 & 0.83 & \textbf{0.87} & 0.85 & 0.82 & 0.83 & 0.82 & 0.89 & 0.77 & 0.82 & \textbf{0.92} & 0.80 & \textbf{0.86}          \\
Standing & 0.85 & 0.78 & 0.81 & \textbf{0.89} & 0.81 & 0.85 & 0.82 & 0.82 & 0.82 & 0.82 & 0.81 & 0.81 & 0.83 & \textbf{0.92} & \textbf{0.87}         \\
Walking & 0.88 & 0.95 & 0.91 & 0.90 & 0.94 & 0.92 & \textbf{0.95} & 0.93 & 0.94 & 0.91 & 0.94 & 0.92 & \textbf{0.95} & \textbf{0.96} & \textbf{0.95}         \\ 
Ascending stairs & 0.82 & 0.73 & 0.77 & 0.82 & 0.61 & 0.70 & \textbf{0.84} & 0.70 & 0.77 & 0.80 & \textbf{0.80} & \textbf{0.80} & \textbf{0.84} & 0.73 & 0.78      \\
Descending stairs & 0.85 & 0.69 & 0.76 & 0.80 & 0.73 & 0.77 & 0.88 & \textbf{0.76} & 0.81 & 0.83 & 0.71 & 0.77 & \textbf{0.97} & \textbf{0.76} & \textbf{0.85}         \\
Vacuum cleaning & 0.73 & 0.85 & 0.79 & \textbf{0.74} & 0.89 & \textbf{0.81} & 0.69 & 0.82 & 0.75 & 0.70 & 0.88 & 0.78 & 0.72 & \textbf{0.92} & \textbf{0.81}            \\
Ironing & 0.91 & 0.93 & 0.92 & 0.93 & \textbf{0.97} & 0.95 & 0.93 & \textbf{0.97} & 0.95 & \textbf{0.94} & 0.96 & 0.95 & \textbf{0.94} & \textbf{0.97} & 0.95     \\
\hline
Average & 0.86 & 0.84 & 0.85 & 0.86 & 0.85 & 0.85 & 0.87 & 0.85 & 0.85 & 0.86 & 0.85 & 0.85 & \textbf{0.90} & \textbf{0.88} & \textbf{0.88} \\
\bottomrule
\end{tabular}
}
\vspace{-.2in}
\end{table}

\begin{figure}[t!]
	\centering
	\subfigure[ERM]{
		\label{fig-cm-erm} 
		\includegraphics[width=.32\linewidth]{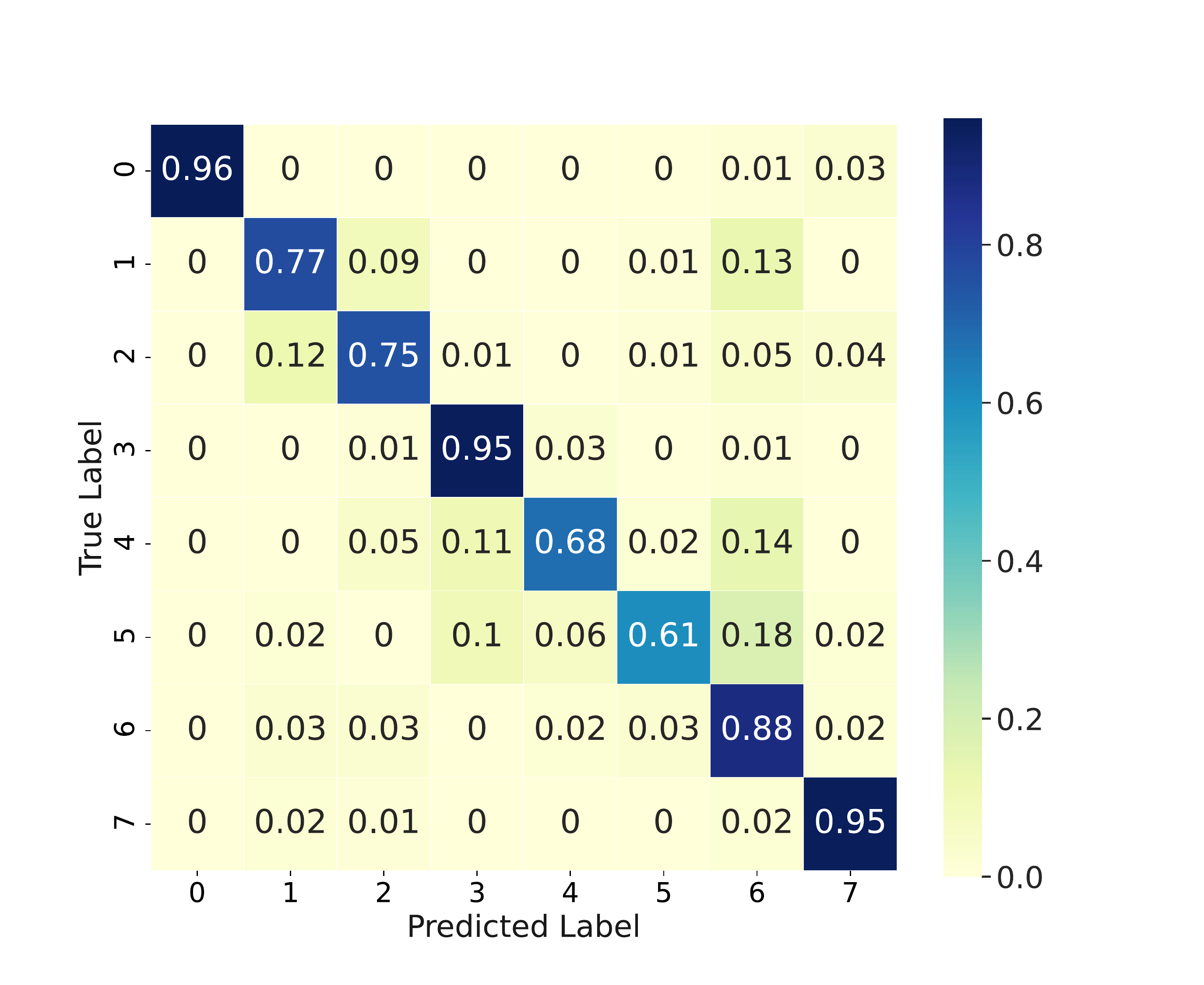}}
	\subfigure[DANN]{
		\label{fig-cm-dann} 
		\includegraphics[width=.32\linewidth]{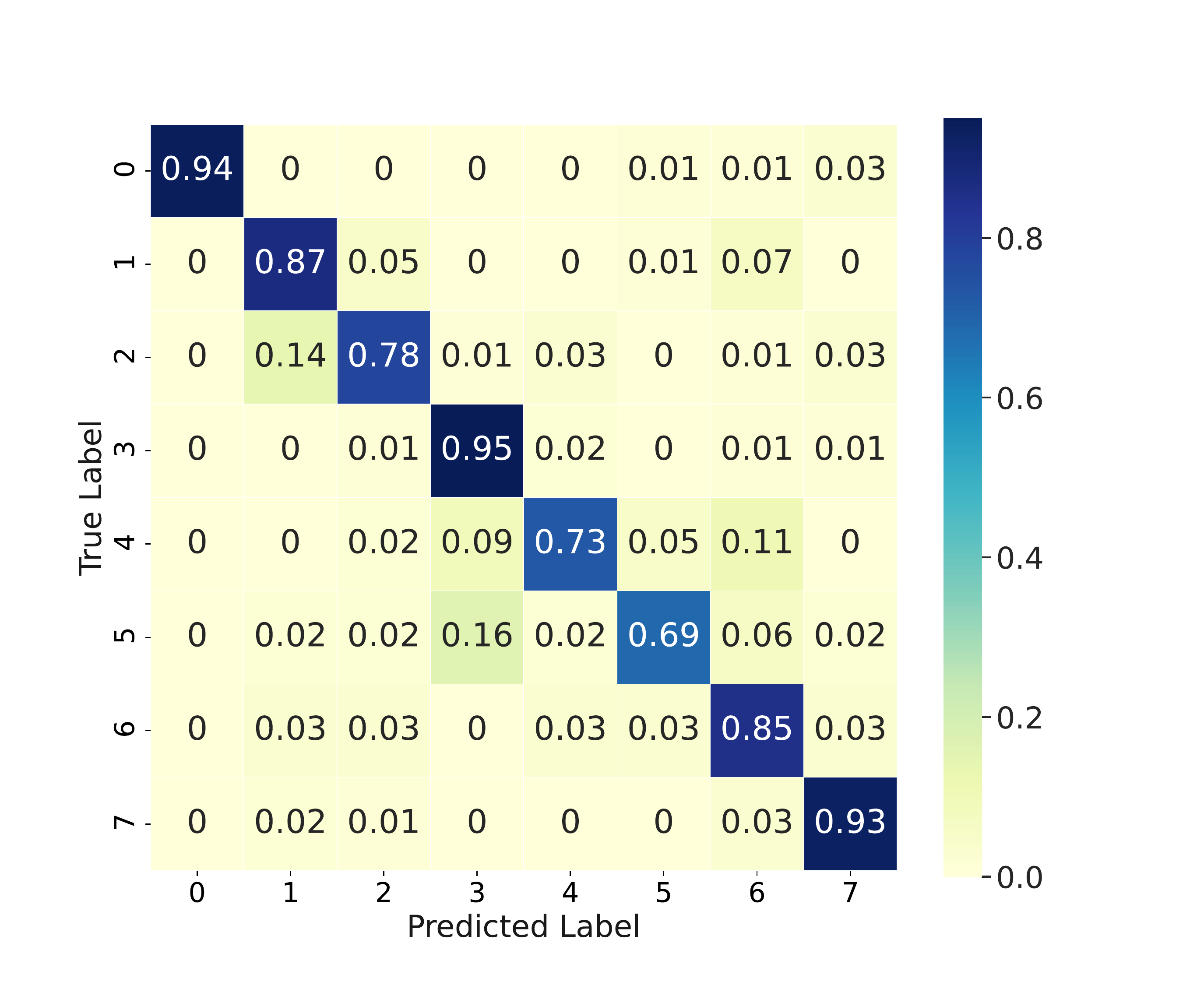}}
	\subfigure[GroupDRO]{
		\label{fig-cm-group} 
		\includegraphics[width=.32\linewidth]{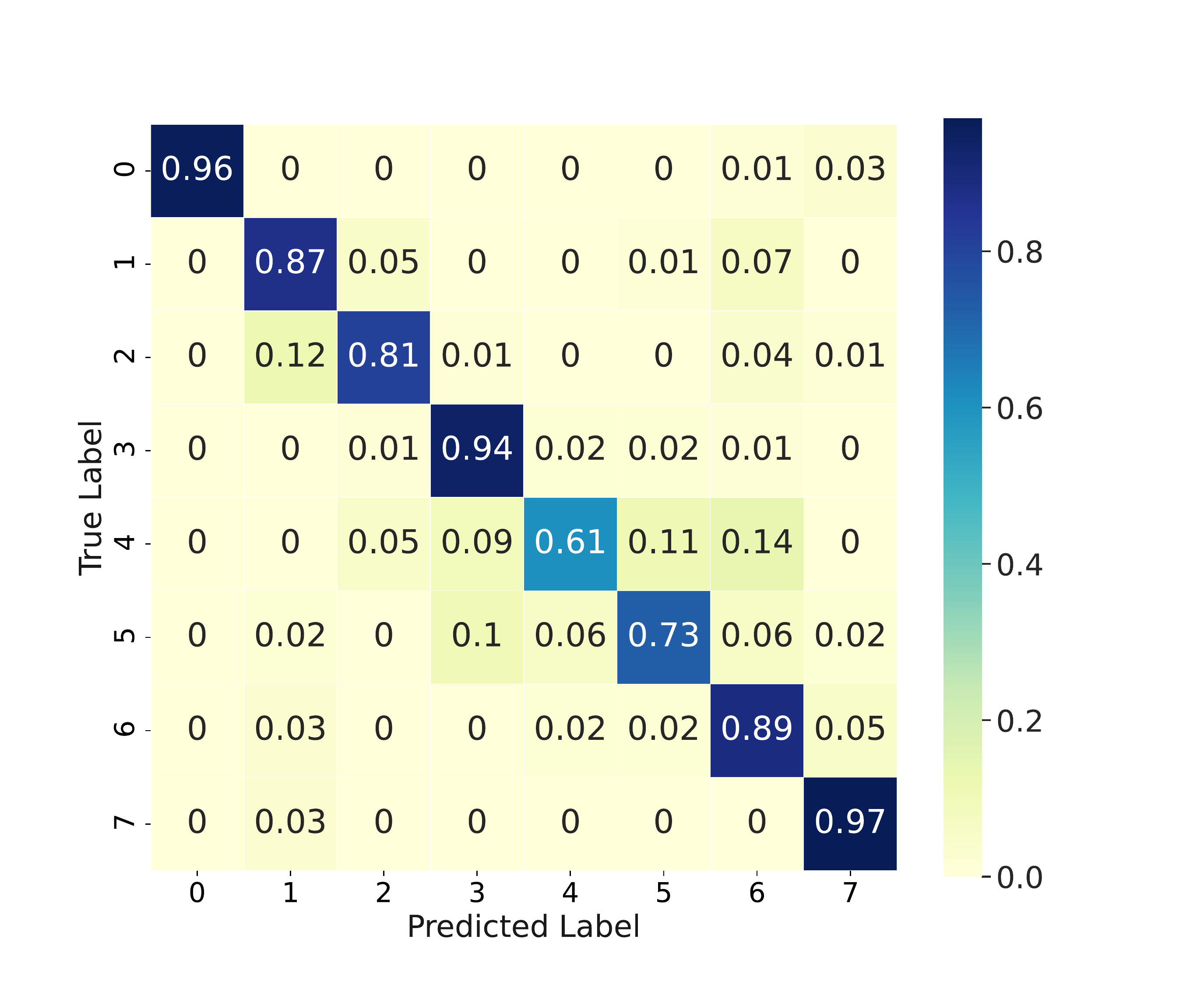}}
	\subfigure[RSC]{
		\label{fig-cm-rsc} 
		\includegraphics[width=.32\linewidth]{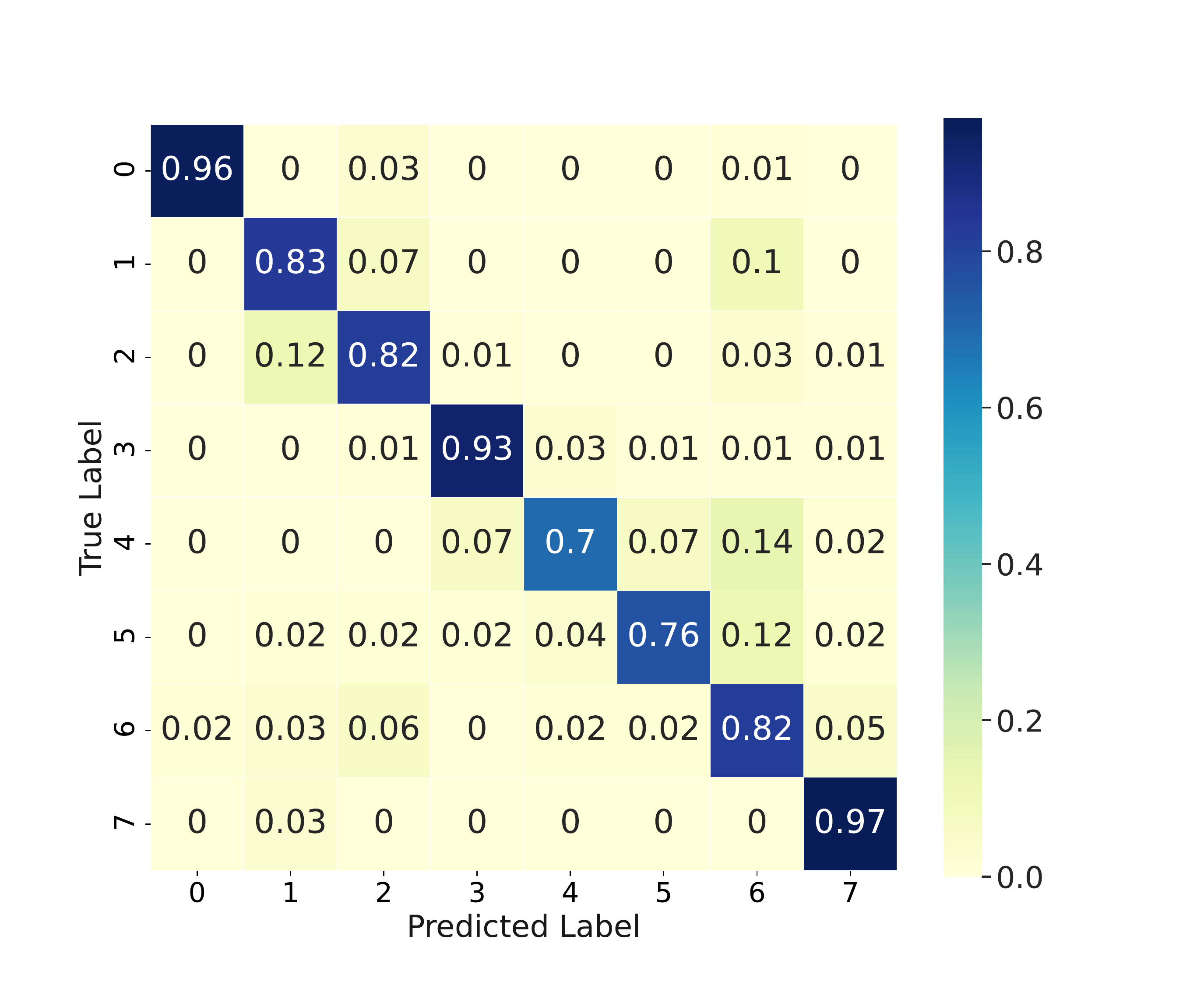}}
	\subfigure[AND-mask]{
		\label{fig-cm-andmask}
		\includegraphics[width=.32\linewidth]{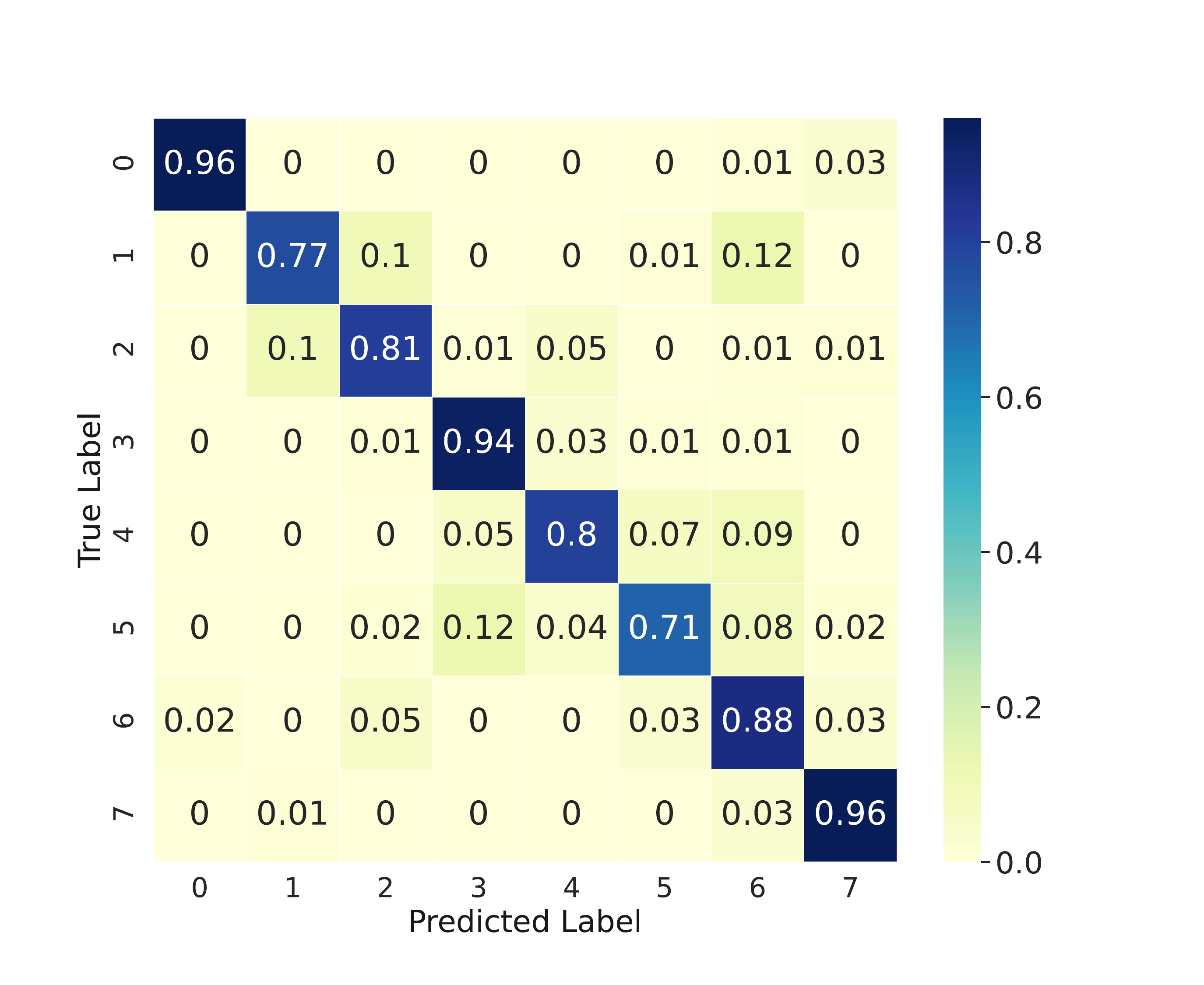}}
	\subfigure[Ours]{
		\label{fig-cm-aff} 
		\includegraphics[width=.32\linewidth]{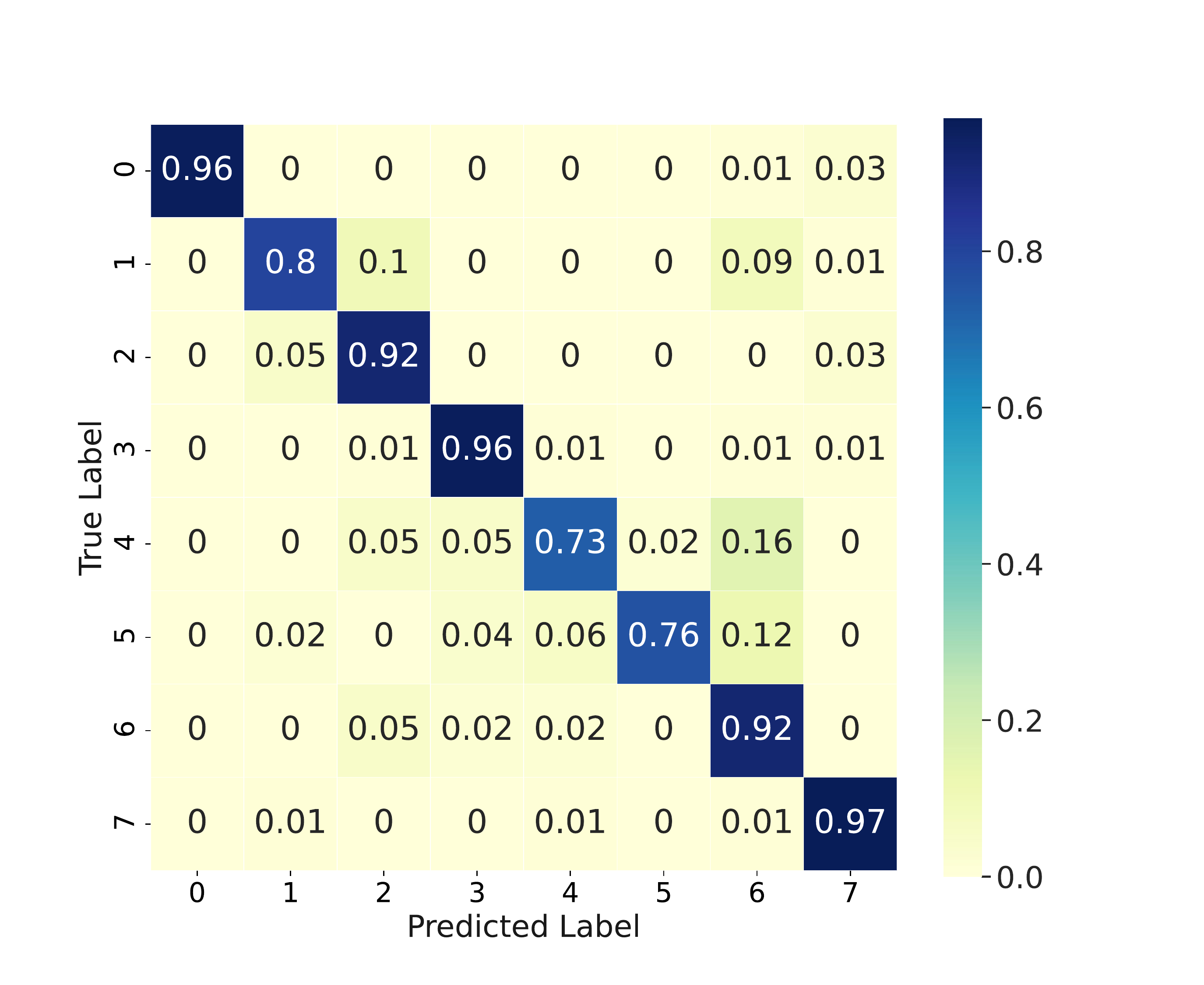}}
	\vspace{-.1in}
	\caption{Confusion matrices of all methods on PAMAP2. Classes 0-7 denote the following activities: lying, sitting, standing, walking, ascending stairs, descending stairs, vacuum cleaning, and ironing.}
	\label{fig-confusion matrix} 
	\vspace{-.2in}
\end{figure}

Meanwhile, we randomly select a task of PAMAP2 to further make the fine-grained evaluation from the perspective of comparison with some state-of-the-art methods by utilizing the metrics of multi-class precision (P), recall (R), F1 scores (F1) in \tablename~\ref{tb-prf-pamap-t1} and the visualization of the confusion matrix for each category in \figurename~\ref{fig-confusion matrix}. From these experimental results, it can be observed that \method can get the most number of the best precision-recall-F1 results on all categories than other comparison methods, and can get the best average precision-recall-F1 across each class. Besides, from the confusion matrix, we can see that \method can get more balance results on each category. Most methods get less satisfying results on the fourth and fifth class, \method can reduce the performance degradation. Although AND-mask can get the best performance on the fourth class, it gets less satisfying results on other classes than \method except for the first class where all the comparative methods can achieve a satisfying performance.

\subsection{Parameter Sensitivity, Convergence, and Time Analysis}

\begin{figure}[t!]
    \centering
    \vspace{-.1in}
    \subfigure[Sensitivity of $\lambda$]{
    \includegraphics[width=.23\textwidth]{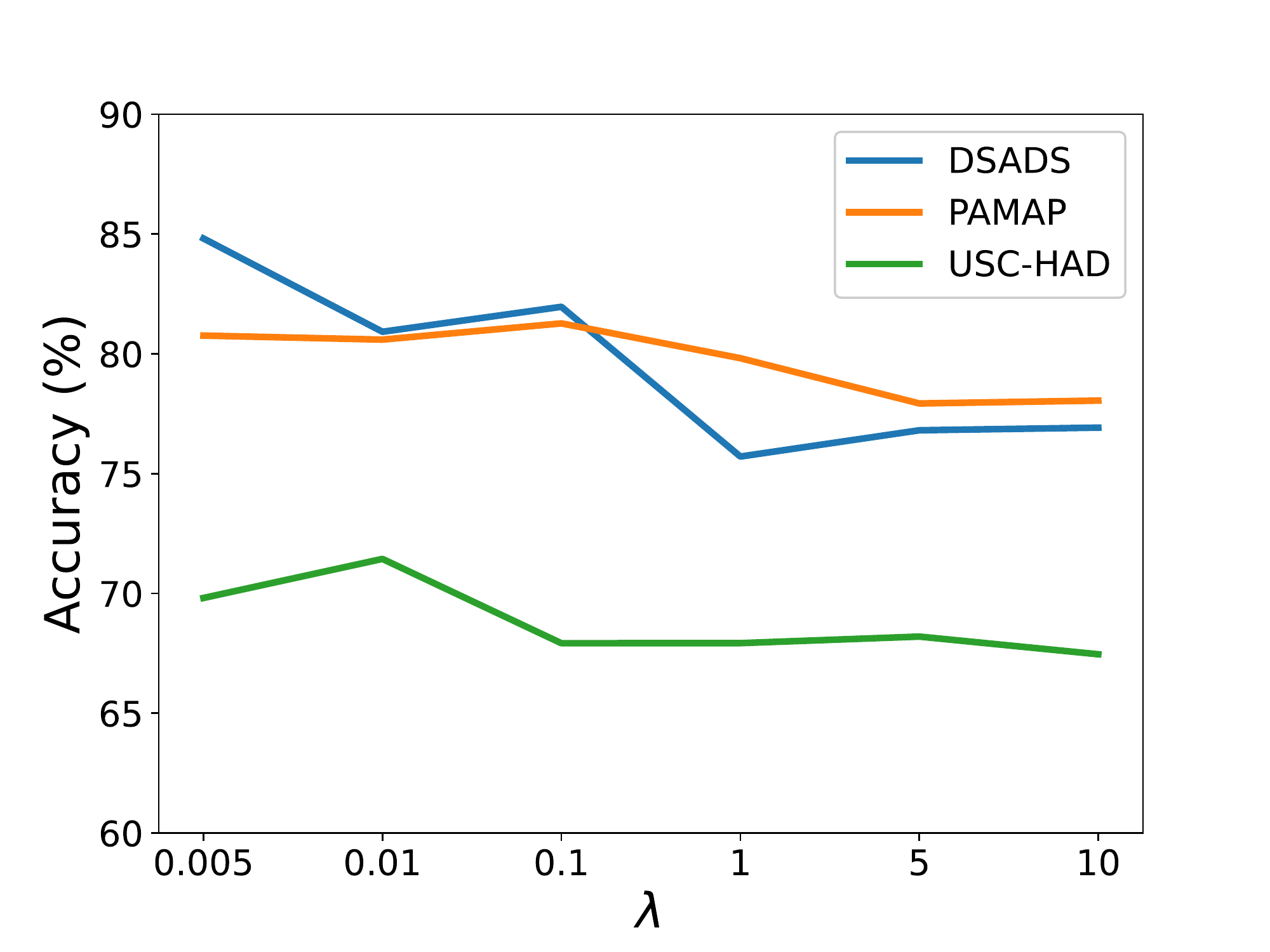}
    \label{fig-lambda}
    }
    \subfigure[Sensitivity of $\beta$]{
    \includegraphics[width=.23\textwidth]{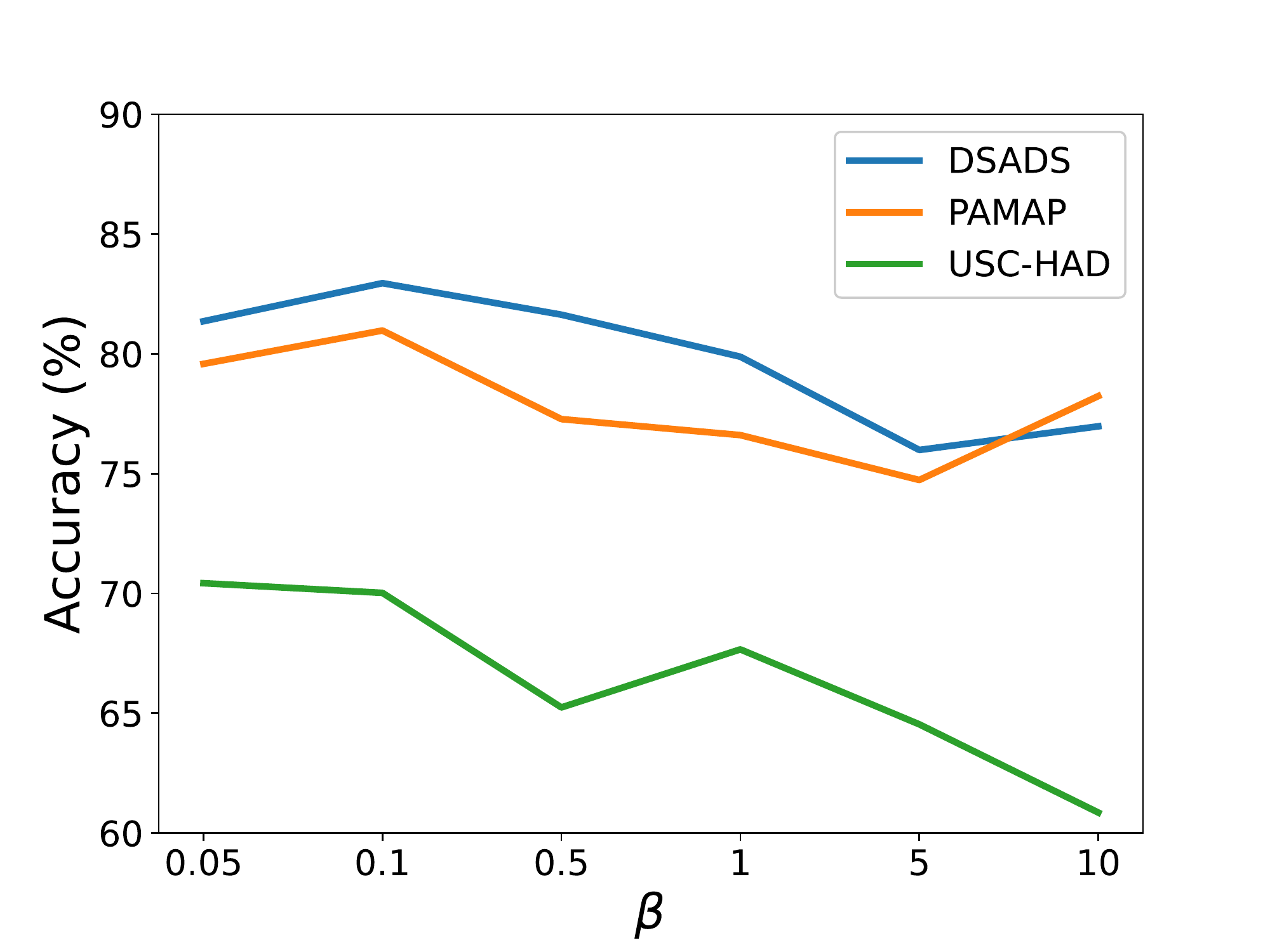}
    \label{fig-beta}
    }
    \subfigure[Convergence]{
    \includegraphics[width=.23\textwidth]{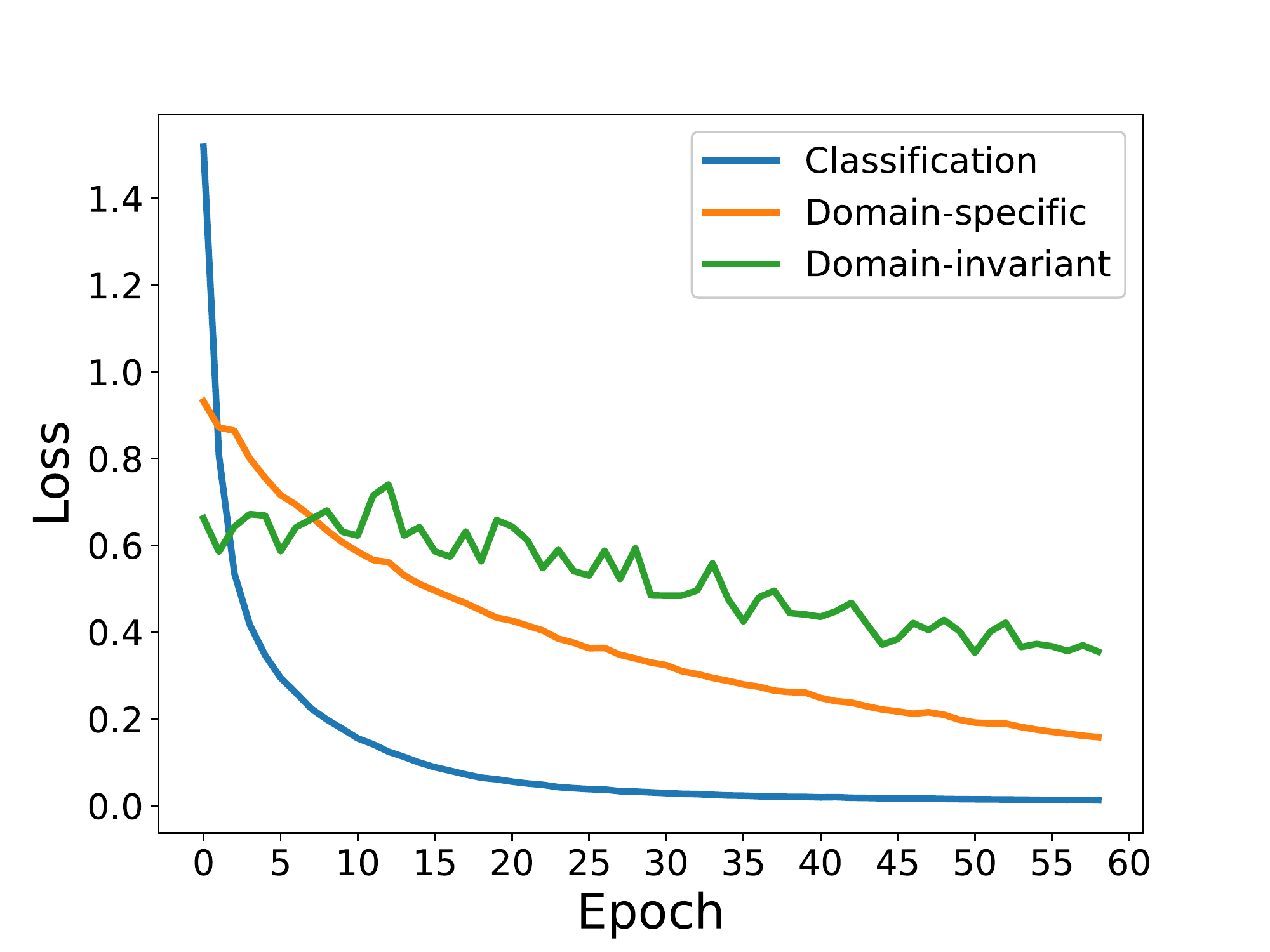}
    \label{fig-conv}
    }
    \subfigure[training Time]{
    \includegraphics[width=.21\textwidth]{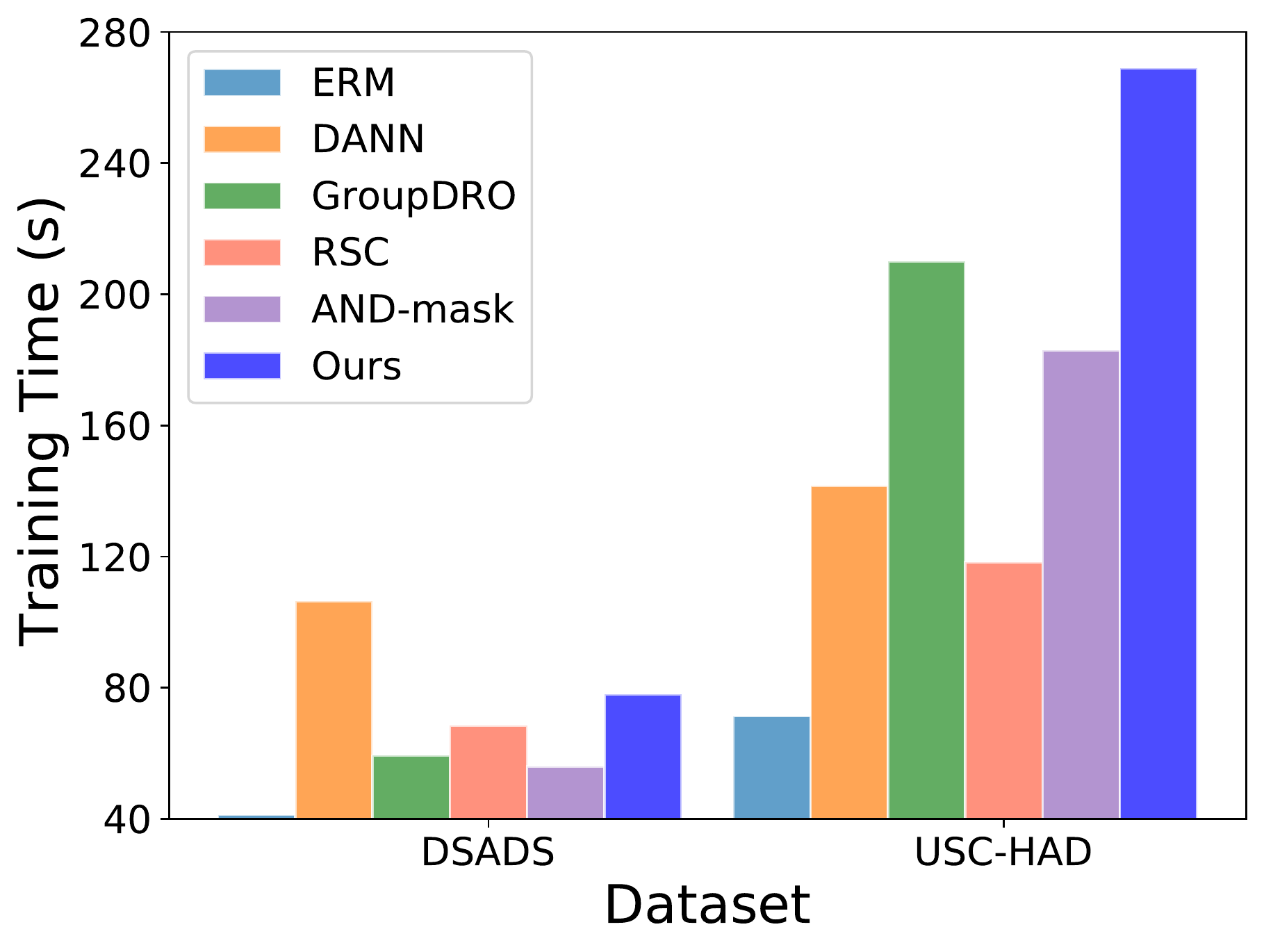}
    \label{fig-time}
    }
    \vspace{-.1in}
    \caption{Parameter sensitivity of (a) $\lambda$ and (b) $\beta$. (c) is convergence analysis. (d) is training time analysis.}
    \vspace{-.2in}
    \label{fig-para}
\end{figure}

We empirically evaluate the sensitivity of two parameters $\lambda$ and $\beta$ by setting their values from $\{0.005, 0.01, 0.1, 1, 5, 10\}$ and $\{0.05, 0.1, 0.5, 1, 5, 10\}$, respectively .
The results are shown in \figurename~\ref{fig-lambda} and \ref{fig-beta}.
It indicates that our \method stays robust to a wide range of parameter choices.

We also empirically analyze the convergence of our method and draw the loss curve on a randomly chosen task of the PAMAP2 dataset in \figurename~\ref{fig-conv}. Other tasks follow similar observations.
The results show that \method can converge in dozens of epochs, indicating that it is easy to train.

Furthermore, we show the training time of each method on one task in \figurename~\ref{fig-time}. We see that the training time of our method is almost the same as others, while slightly takes more time in some circumstances. This is reasonable because our method is an ensemble-based learning process with several domain-specific branches to be learned, which is comparable with other methods. We obtain similar observations for the inference time.

\begin{table}[t!]
\caption{Comparison of inference time and weighted F1-score}

\begin{tabular}{l|cccccc}
\toprule
Method             & ERM      & DANN     & GroupDRO & RSC      & AND-mask & \method (Ours)     \\ \midrule
Inference time (s) & .000010 & .000012 & .000011 & .000011 & .000012 & .000057 \\
Weighted F1 (\%)   & 82.10     & 82.90     & 82.07    & 82.15    & 81.68    & 85.43    \\ \bottomrule
\end{tabular}
\label{tb-infer-time}
\end{table}

As inference time is also very crucial to make quick classification while performing accurate activity recognition. Table~\ref{tb-infer-time} shows the comparison of the average inference time of each sample and weighted F1 score on the DSADS dataset. We can observe that the inference time of different methods are similar and the proposed method is slightly longer, while the recognition performance is the best among the comparison methods. This indicates the method can make better activity recognition in applications with a little more time. And we will make efforts to further reduce the inference time for better applications in future work.

\section{Application to ADHD Recognition}
\label{sec-app}

Attention Deficit Hyperactivity Disorder~(ADHD) is one of the most common mental disorders in children~\cite{jiang2020weda}. ADHD is characterized by inappropriate inattention, hyperactivity, and impulsivity~\cite{barkley1998attention}. It is often accompanied by some motor abnormality, thus it is possible and necessary to utilize the HAR method to assist the diagnosis. In this section, we apply the proposed \method to the ADHD application to further evaluate its effectiveness.

We apply our algorithm to a real-world ADHD dataset~\cite{jiang2020weda}. This dataset is collected with a designed wearable diagnostic assessment system in the room environment with little tension. Ten diagnostic tasks including six interactions with the screen tasks and four physical objects interaction tasks (Schulte grid, Multi-ball tracking, Catching grasshopper, Drinking birds, Limb reaction, Reading, Finger holes, Shape-color conflicting, Catching worms, and Keeping balance) are designed for children for ADHD symptoms according to DSM-5. Six wearable sensors are attached to children, one on each wrist and each ankle, one on the head, one on the waist during the ten assessment tasks and accelerometer data are collected. 54-dimensional motion features are extracted according to \cite{jiang2020weda}. For more detailed information about this dataset, please refer to \cite{jiang2020weda}.

\begin{wrapfigure}{r}{.5\textwidth}
\centering
\vspace{-.2in}
\includegraphics[width=.5\textwidth]{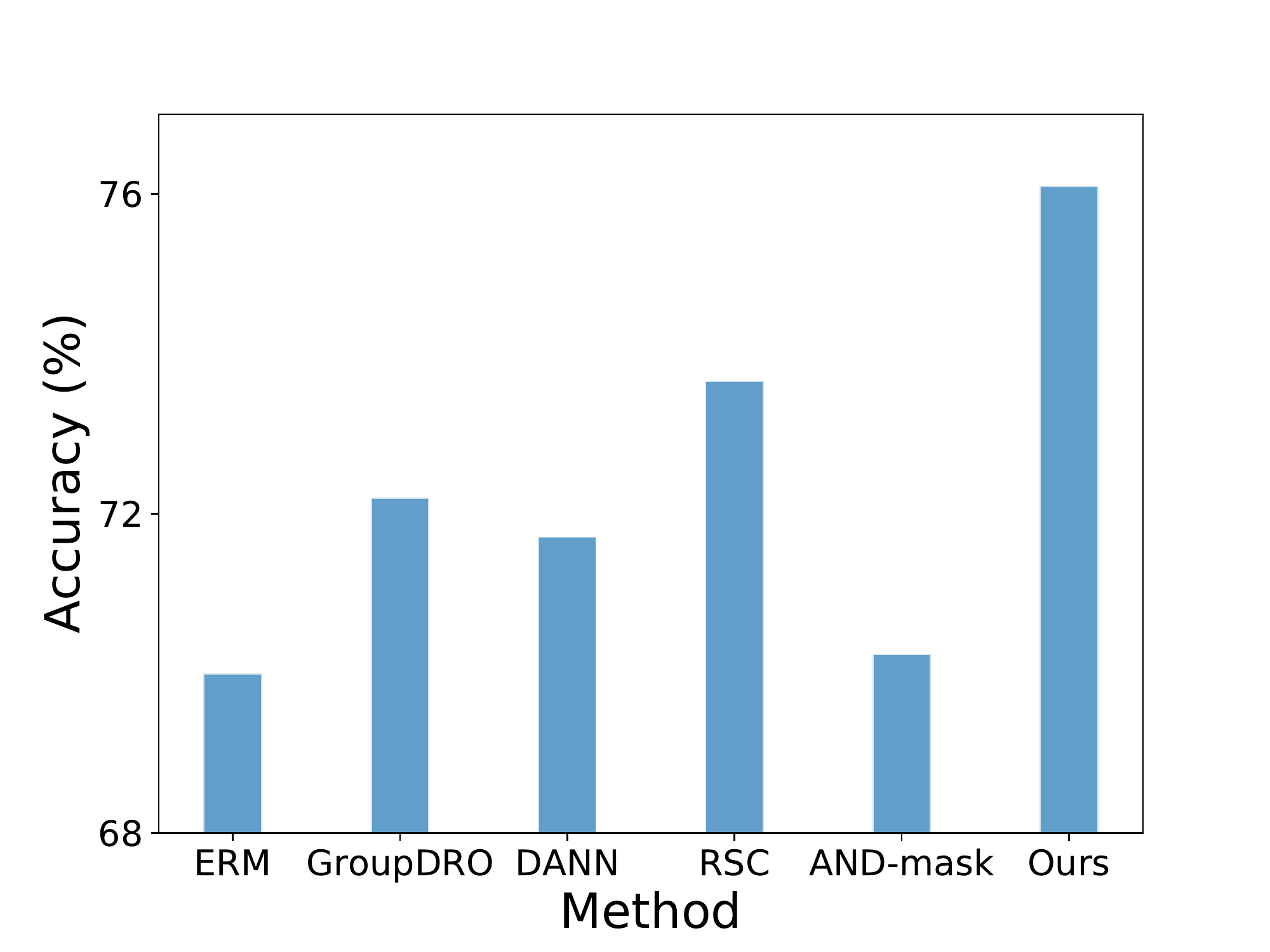}
  \vspace{-.1in}
  \caption{Multi-class classification in ADHD application.}
  \label{fig-adhd-multi}
\end{wrapfigure}

In our experiment, features from 83 normal children and 83 children with ADHD are involved. Meanwhile, the children with ADHD are diagnosed by doctors to confirm they meet the diagnostic criteria for ADHD. Furthermore, we have more fine-grained labels, i.e., children with ADHD are diagnosed with subtypes, i.e. predominantly inattentive~(50 children), predominantly hyperactive-impulsive~(14 children), and combination~(19 children).
This is a multi-class classification task.

We divide the data into the source training set and target test set, with a number of participants of 125 and 41, respectively.
To harness the different distributions in the training set, we further divide it into 3 domains.
The classification results of all methods are shown in \figurename~\ref{fig-adhd-multi}.
We can observe that \method improves the classification accuracy with a rate of around 2.44\%.
It indicates the effectiveness of \method in classification with ADHD and the effectiveness of fine-grained classification on subtypes. Experimental results also show that \method has the potential to the applied in real-life wearable healthcare.

\section{Conclusions and Future Work}
\label{sec-con}

Generalization to the unseen test data has always been the key research and application problem in human activity recognition. While transfer learning and domain adaptation approaches rely on the availability of test data during the training stage, in this paper, we propose \method to solve this problem by learning both the domain-invariant and domain-specific features. The key of our algorithm is to preserve the specific representations of the training data while learning transferable representations, which could be informative to the generalization on unseen test data.
Experiments on both public datasets and the real application have demonstrated the superiority of our method.

In the future, we plan to extend \method in the following two directions. First, apply it to more healthcare applications such as the diagnosis of Parkinson's disease. Second, experiments show that there is still a gap between our method and fine-tuning, which motivates us to improve the performance of our method by introducing other domain-invariant learning modules.

\begin{acks}
This work is supported by National Key Research and Development Plan of China (No. 2021YFC2501202), Natural Science Foundation of China (No. 61972383, No. 61902377, No. 62101530, No. 61902379), Science and Technology Service Network Initiative, Chinese Academy of Sciences (No. KFJ-STS-QYZD-2021-11-001).
\end{acks}

\bibliographystyle{ACM-Reference-Format}
\bibliography{refs}

\end{document}